\documentclass[11pt,a4paper]{article}
\usepackage[hyperref]{emnlp2020}
\usepackage{times}
\usepackage{latexsym}

\usepackage{graphicx}
\usepackage{multirow}
\usepackage{url}
\usepackage{amssymb}
\usepackage{amsmath}
\usepackage{enumitem}
\usepackage{color,soul}
\usepackage{mathtools}
\usepackage[
 font=small 
 ]{subfig}

\usepackage{microtype}

\aclfinalcopy 
\def\aclpaperid{1654} 

\title{Probing Task-Oriented Dialogue Representation from Language Models}

\author{Chien-Sheng Wu and Caiming Xiong \\
Salesforce Research \\
\texttt{[wu.jason, cxiong]@salesforce.com}
}

\date{}

\begin{document}
\maketitle
\begin{abstract}
This paper investigates pre-trained language models to find out which model intrinsically carries the most informative representation for task-oriented dialogue tasks.
We approach the problem from two aspects: supervised classifier probe and unsupervised mutual information probe.
We fine-tune a feed-forward layer as the classifier probe on top of a fixed pre-trained language model with annotated labels in a supervised way.
Meanwhile, we propose an unsupervised mutual information probe to evaluate the mutual dependence between a real clustering and a representation clustering.
The goals of this empirical paper are to
1) investigate probing techniques, especially from the unsupervised mutual information aspect,
2) provide guidelines of pre-trained language model selection for the dialogue research community,
3) find insights of pre-training factors for dialogue application that may be the key to success.



\end{abstract}

\section{Introduction}

Task-oriented dialogue systems achieve specific user goals within a limited number of dialogue turns via natural language. They have been used in a wide range of applications, such as booking restaurants~\cite{wen2016network}, providing tourist information~\cite{budzianowski2018multiwoz,wu2019global}, ordering tickets~\cite{schulz2017frame}, and healthcare consultation~\cite{wei2018task}. They are also crucial components of intelligent virtual assistants like Siri, Alexa, and Google Assistant. 

\begin{table}
\centering
\resizebox{\linewidth}{!}{
\begin{tabular}{r|ccc}
\hline
\textbf{Model} & \textbf{Dial. Data} & \textbf{Parameters} & \textbf{Output Dim.} \\ \hline
BERT-base & X & 109.5M & 768 \\
AlBERT-base & X & 11.7M & 768 \\
DistilBERT-base & X & 66.4M & 768 \\
RoBERTa-based & X & 124.6M & 768 \\
GPT2-small & X & 124.4M & 768 \\
ELECTRA-GEN & X & 33.5M & 256 \\
ELECTRA-DIS & X & 108.9M & 768 \\
ConveRT & V & 29M & 1024 \\
DialoGPT-small & V & 124.4M & 768 \\
TOD-BERT-mlm & V & 119.5M & 768 \\
TOD-BERT-jnt & V & 119.5M & 768 \\
TOD-GPT2 & V & 124.4M & 768 \\
\hline
\end{tabular}
}
\caption{An overview of selected pre-trained language models (Details in Section~\ref{sec:pretrainedLM}).}
\label{tab:model}
\end{table}

Most of the task-oriented dialogue systems nowadays, are benefited from transfer learning~\cite{WuTradeDST2019,lin2020mintl}, especially pre-trained language models trained on general text, such as BERT~\cite{devlin2018bert} and GPT2~\cite{radford2019language}. 
However, previous work claims that linguistic patterns could differ between writing text and human conversation, resulting in a large gap of data distributions~\cite{bao2019plato,wolf2019transfertransfo}. Recently, several approaches are leveraging open-domain data~\cite{henderson2019convert,zhang2019dialogpt}, or aggregating task-oriented data~\cite{wu2020tod} to pre-train language models. 

In this paper, we are interested in answering these questions: which language model has the most informative representations that is better for what task-oriented dialogue task? Does pre-training with dialogue-specific data or different objectives make any difference? 
We investigate how good these pre-trained representations are for a task-oriented dialogue system, ignoring the model architectures and training strategies by only probing their final representations with fine-tuning models. A good representation implies better knowledge transferring and domain generalization ability, making downstream applications easier and cheaper to be improved. 

We tackle this problem with two probing solutions: supervised classifier probe and unsupervised mutual information probe.
Classifier probe is commonly used in different NLP tasks such as morphology~\cite{belinkov-etal-2017-neural}, sentence length~\cite{adi2016fine}, or linguistic structure~\cite{hewitt2019structural}. In this setting, we fine-tune a simple classifier for a specific task (e.g., intent identification) on a fixed pre-trained language model. The probe uses supervision to find the best transformation for each sub-task.

In addition, we present mutual information probe to investigate these language models by directly clustering their output representations, as recent study~\cite{pimentel2020information} suggests that a simple classifier may not be able to achieve the best estimate of mutual information between features and the downstream task.
We apply two clustering techniques, K-means~\cite{lloyd1982least} and Gaussian mixture model~\cite{reynolds2009gaussian}, to calculate its adjusted normalized mutual information (ANMI)~\cite{vinh2010information} between the predicted clustering and the true task-specific clustering.

We investigate 12 language models, as shown in Table~\ref{tab:model}, where five of them have been pre-trained with dialogue data.
We evaluate four core task-oriented dialogue tasks, domain identification, intent detection, slot tagging, and dialogue act prediction. They correspond to the commonly defined natural language understanding, dialogue state tracking, and dialogue management modules \cite{wen2016network}.
We hope our probing analysis can provide insights to facilitate future task-oriented dialogue research.
Some of the key observations in this work are summarized here (More discussion in Section~\ref{subsec-results}):
\begin{itemize}[leftmargin=*]
    \item No matter the open-domain or close-domain, pre-training with dialogue data helps learning better representations for task-oriented dialogue. 
    
    \item Pre-trained language models intrinsically contain more information about intents and dialogue acts but less for slots.
    
    \item ConveRT~\cite{henderson2019convert} and TOD-BERT-jnt~\cite{wu2020tod} have the highest classification accuracy and mutual information score, suggesting that response selection is useful for dialogue pre-training, especially when we compare TOD-BERT-jnt to TOD-BERT-mlm.
    
    \item Top models also include TOD-GPT2 and DistilBERT~\cite{sanh2019distilbert}. The distilled version of BERT surprisingly outperforms BERT and other strong baselines such as RoBERTa~\cite{liu2019roberta}.
    
    \item DialoGPT and GPT2 do not perform well on mutual information evaluation but have a middle-ranking classification accuracy, implying that their representations are informative but not suitable for unsupervised clustering.
    
    \item Models such as AlBERT~\cite{lan2019albert} and ELECTRA~\cite{clark2020electra} have low classification accuracy and mutual information, showing the least useful information on task-oriented dialogue tasks. 
\end{itemize}

\section{Pre-Trained Language Models}
\label{sec:pretrainedLM}
W can roughly divide pre-trained language models into two categories: uni-directional and bi-directional. BERT-based systems are bi-directional language models and usually trained with the masked language modeling (MLM) objective, i.e., given the left and right context to predict the current masked token. 
GPT-based models, on the other hand, are uni-directional language models trained always to predict the next token in an auto-regressive way. 

For a BERT-based model, we use the final-layer hidden state of its first token, [CLS], to represent an input sequence. This built-in token is originally designed to aggregate the information. Since GPT-based models are uni-directional and do not have a similar design as the [CLS] token, we use the mean pooling of its output hidden states to represent the input sequence, which is better than only using the last hidden state in our experiments.

\paragraph{BERT-based} BERT is a Transformer~\cite{vaswani2017attention} encoder with a self-attention mechanism, which is trained on Wikipedia and BookCorpus using the MLM and next sentence prediction objectives. \citet{liu2019roberta} proposed a robustly optimized approach for BERT, call RoBERTa, where they improved it by training the model longer with bigger batches over more data and longer sequences, and removing the next sentence prediction objective. \citet{lan2019albert} proposed a lite BERT (AlBERT) that trained with MLM and inter-sentence coherence losses, and aimed to lower memory consumption and increase the training speed. With similar motivation, \citet{sanh2019distilbert} trained a DistilBERT that reduce 40\% of parameters with a triple loss, including MLM, distillation, and cosine-distance losses. \citet{clark2020electra} proposed ELECTRA using a sample-efficient pre-training task called replaced token detection. They used a generator network (ELECTRA-GEN) to replace tokens with plausible alternative tokens and trained a discriminative model (ELECTRA-DIS) to predict whether the generator replaced each token in the input.

Most of the pre-trained models above are trained on general text corpora with language modeling objectives. \citet{henderson2019convert}, on the other hand, used social media conversational data to train the ConveRT model. It is a Transformer-based dual-encoder model pre-trained on a dialogue response selection task using 727M Reddit (input, response) pairs. Very recently, \citet{wu2020tod} proposed task-oriented dialogue BERT (TOD-BERT), which is initialized by BERT and further pre-trained on nine publicly available task-oriented dialogue corpora. They have one version with only MLM objective (TOD-BERT-mlm) and another with both MLM and contrastive learning objectives of response selection (TOD-BERT-jnt). TOD-BERT has shown good performance on several task-oriented downstream tasks, especially in the few-shot setting.

\paragraph{GPT-based} GPT2 \cite{radford2019language} is the representative of uni-directional language models using a Transformer decoder, where the objective is to maximize left-to-right generation likelihood. To ensure diverse and nearly unlimited text sources, they use Common Crawl to obtain 8M documents as its training data. 
\citet{budzianowski2019hello} trained GPT2 on task-oriented response generation task, taking system belief, database result, and last dialogue turn as inputs. It only uses one dataset to train its model because few public datasets have database information available for pre-training.
\citet{zhang2019dialogpt} pre-trained GPT2 on 147M open-domain Reddit data for response generation and called it DialoGPT. It aims to generate more relevant, contentful, and consistent responses for chit-chat dialogue systems. In this paper, following TOD-BERT's idea, we train a task-oriented GPT2 model (TOD-GPT2) built on the GPT2 model and further pre-trained with task-oriented datasets. We use the same dataset collection, which contains nine datasets in total, as shown in \citet{wu2020tod}, to pre-train the model as a reference.

\section{Method}
We define a dialogue corpus $D = \{D_1,\dots,D_M\}$ has $M$ dialogue samples, and each dialogue sample $D_m$ has $T$ turns of conversational exchange $\{U_1,S_1\dots,U_T,S_T\}$ between a user and a system. For every utterance $U_t$ or $S_t$, we have human-annotated domain, user intent, slot, and dialogue act labels. 
We first feed all the utterances to a pre-trained model and obtain user and system representations.
In this section, we first discuss how we design our classifier probe and then introduce our mutual information probe's background and usage. 

\subsection{Classifier Probe}
We use a simple classifier to transform those representations for a specific task and optimize it with annotated data. 
\begin{equation}
\begin{array}{c}
    V_i = A(FFN(E_i)),
\end{array}
\end{equation}
where $E_i \in \mathbb{R}^{d_B}$ is the output representation with dimension $d_B$ from a pre-trained model, $FFN \in \mathbb{R}^{N \times d_B}$ is a feed-forward layer that maps from dimension $d_B$ to a prediction with $N$ classes, and $A$ is an activation layer. For domain identification and intent detection, we use a Softmax layer and backpropagate with the cross-entropy loss. For dialogue slot and act prediction, we use a Sigmoid layer and the binary cross-entropy loss since they are multi-label classification tasks.

\subsection{Mutual Information Probe}

We first cluster utterances in an unsupervised fashion using either K-means~\cite{lloyd1982least} or Gaussian mixture model (GMM)~\cite{reynolds2009gaussian} with $K$ clusters. Then we compute the adjusted mutual information score~\cite{vinh2010information} between the predicted clustering and each of the true clusterings (e.g., domain and intent) for different hyper-parameters $K$. Note that the predicted clustering is not dependent on any particular labels.

\subsubsection{Utterance Clustering}
K-means is a common clustering algorithm that aims to partition $N$ samples into $K$ clusters $A = \{A_1,\dots,A_K\}$ in which each sample is assigned to a cluster centroid with the nearest mean. 
\begin{equation}
\begin{array}{c}
    \text{arg}\,\max\limits_{A} \sum\limits_{i=1}^{K}\sum\limits_{x \in A_i} \Vert x-\mu_i \Vert^2,
\end{array}
\end{equation}
where $\mu_i$ is the centroid of the $A_i$ cluster and the algorithm is updated in an iterative manner. 

On the other hand, GMM assumes a certain number of Gaussian distributions ($K$ mixture components). It takes both mean and variance of the data into account, while K-means only consider the data's mean. By the Expectation-Maximization algorithm, GMM first calculates each sample's probability belongs to a cluster $A_i$ during the E-step, then updates its density function to compute new mean and variance during the M-step.

In our experiments, we cluster separately for user utterances $U$ and system response $S$. 
Note that $K$ is a hyper-parameter since we may not know the true distribution in a real scenario. 
To avoid the local minimum issue, we run multiple times (typically ten runs) and use the best clustering result for mutual information evaluation.

\subsubsection{ANMI}
To evaluate two clusterings' quality, we compute the ANMI score between a clustering and its ground-truth annotation. ANMI is adjusted for randomness, which accounts for the bias in mutual information, giving high values to the clustering with a larger number of clusters. ANMI has a value of 1 when two partitions are identical, and an expected value of 0 for random (independent) partitions.

More specifically, we assume two label clusterings, $A$ and $B$, that have the same $N$ objects. The mutual information (MI) between $A$ and $B$ is defined by
\begin{equation}
\begin{array}{c}
    \textnormal{MI}(A, B) = \sum\limits_{i=1}^{|A|} \sum\limits_{j=1}^{|B|}  P(i, j) \log(\frac{P(i, j)}{P(i)P(j)}),
\end{array}
\end{equation}
where $P(i,j) = |A_i \cap B_j|/N$ is the probability that a randomly picked sample falls into both $A_i$ and $B_j$ classes. Similarly, $P(i) = |A_i|/N$ and $P(j) = |B_j|/N$ are the probabilities that the sample falls into either the $A_i$ or $B_j$ class.

The normalized mutual information (NMI) normalizes MI with the mean of entropy, which is defined as 
\begin{equation}
\begin{array}{c}
    \textnormal{NMI}(A, B) = \frac{\textnormal{MI}(A, B)}{mean(H(A), H(B))},
\end{array}
\end{equation}
where $H(A) = -\sum_{i=1}^{|A|} P(i) \log(P(i))$ is the entropy of the $A$ clustering, which measures the amount of uncertainty for the partition set. 

MI and NMI are not adjusted for chance and will tend to increase as the number of cluster increases, regardless of the actual amount of ``mutual informaiton'' between the label assignments. Therefore, adjusted normalized mutual information (ANMI) is designed to modify NMI score with its expectation, which is defined by
\begin{equation}
\begin{array}{c}
    \textnormal{ANMI} = \frac{\textnormal{MI} - \mathop{\mathbb{E}}[\textnormal{MI}]}{mean(H(A), H(B))- \mathop{\mathbb{E}}[\textnormal{MI}]},
\end{array}
\end{equation}
where the expectation $\mathop{\mathbb{E}}[\textnormal{MI}]$ can be calculated using the equation in \citet{vinh2010information}. 

\begin{table}[t]
\centering
\resizebox{0.7\linewidth}{!}{
\begin{tabular}{|c|c|c|}
\hline
\multicolumn{3}{|c|}{MWOZ} \\ \hline
\textbf{Domain} & \textbf{Dialogue Act} & \textbf{Slot} \\ \hline
\begin{tabular}[c]{@{}c@{}}restaurant\\ hotel\\ attraction\\ train\\ taxi\end{tabular} & \begin{tabular}[c]{@{}c@{}}nobook \\ bye\\ request\\ recommend\\ welcome\\ book\\ greet\\ nooffer\\ reqmore\\ offerbooked\\ select\\ inform\\ offerbook\end{tabular} & \begin{tabular}[c]{@{}c@{}}type\\ book day\\ book people\\ day\\ pricerange\\ leaveat\\ arriveby\\ parking\\ book time\\ name\\ destination\\ internet\\ stars\\ book stay\\ departure\\ area\\ food\\ department\end{tabular} \\ \hline
\end{tabular}
}
\caption{Labels classes in the MWOZ Data.}
\label{tab:mwoz}
\end{table}

\section{Experiments}

\subsection{Datasets}
The multi-domain Wizard-of-Oz (MWOZ) dataset~\cite{budzianowski2018multiwoz} is one of the most common benchmark datasets for task-oriented dialogue systems. We use MWOZ to evaluate domain identification, dialogue slot tagging, and dialogue act prediction tasks. It contains 8420/1000/1000 dialogues for training, validation, and testing sets, respectively. There are seven domains in the training set and five domains in the others. 
There are 13 unique system dialogue acts and 18 unique slots as shown in Table~\ref{tab:mwoz}.

Besides, we use the out-of-scope intent (OOS) dataset~\cite{larson2019evaluation} for our intent detection experiment. The OOS dataset is one of the largest annotated intent datasets, including 15,100/3,100/5,500 samples for the train, validation, and test sets, respectively. It has 150 intent classes over ten domains and an additional out-of-scope intent class, a user utterance that does not fall into any of the predefined intents. The whole intent list is shown in the Appendix.

\subsection{Training Details}
We first process user utterance and system response using the tokenizer corresponding to each per-trained model. To obtain each representation, we run most of the pre-trained models using the HuggingFace~\cite{Wolf2019HuggingFacesTS} library, except the ConveRT~\footnote{\url{https://github.com/PolyAI-LDN/polyai-models}} and TOD-BERT~\footnote{\url{https://github.com/jasonwu0731/ToD-BERT}}. We fine-tune GPT2 using its default hyper-parameters and the same nine datasets as shown in \citet{wu2020tod} to train for TOD-GPT2 model.
For classifier probing, we fine-tune the top layer with a consistent
hyper-parameter setting. We apply AdamW~\cite{loshchilov2017decoupled} optimizer with a learning rate $5e^{-5}$ and gradient clipping 1.0.
We use $K = 4, 8, 16, 32, 64, 128, 256$ with 50 iterations each, and report the moving trend for MI probing. We use GMM clustering from the scikit-learn library, and we adopt the K-means implementation from the faiss library~\cite{JDH17}.
Experiments were conducted on a single NVIDIA Tesla V100 GPU.

\begin{figure}[!t]
    \centering
    \subfloat[MWOZ Domain (System)]{
        \includegraphics[width=0.9\linewidth]{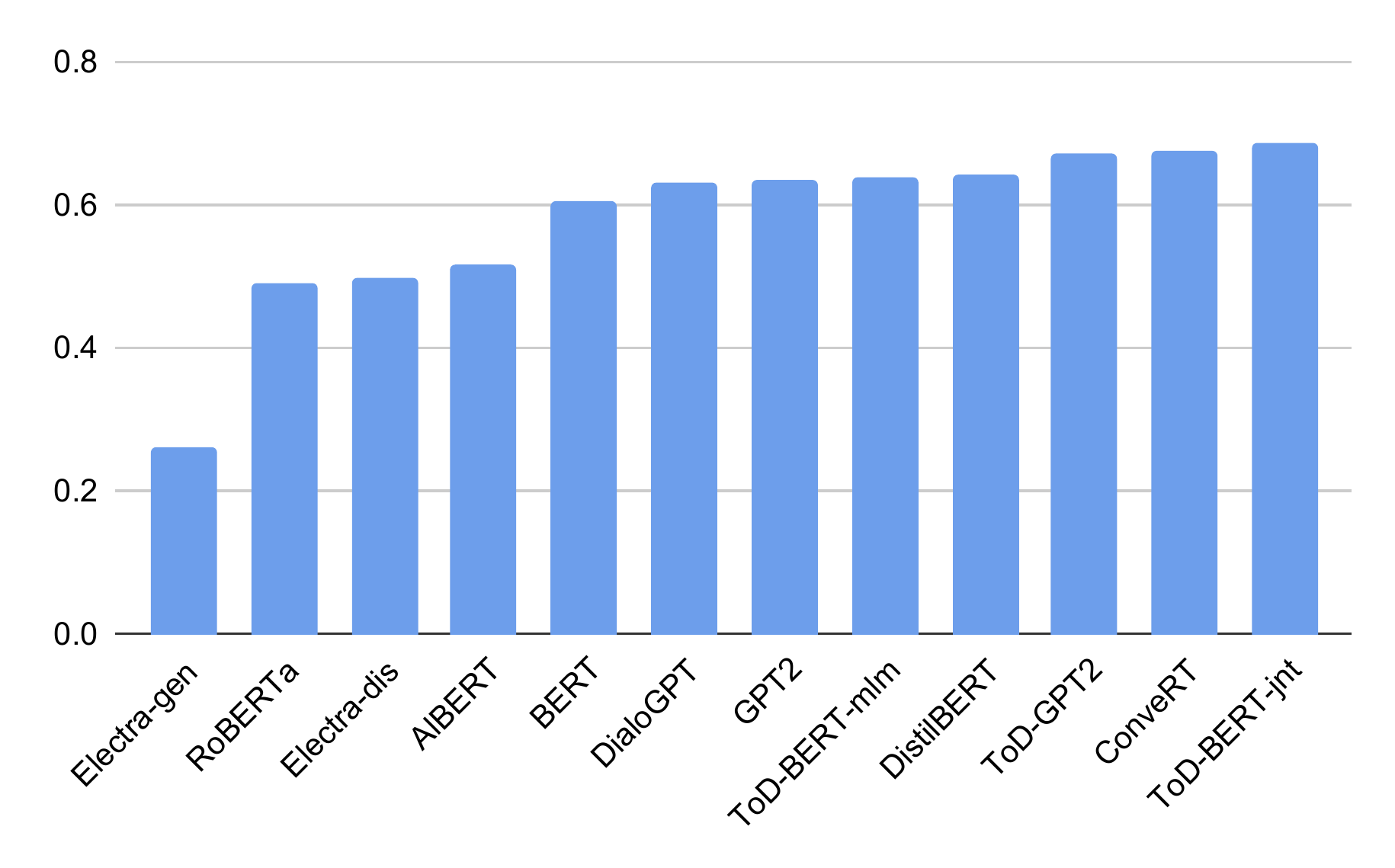}} \\ 
    \subfloat[OOS Intent (User)]{
        \includegraphics[width=0.9\linewidth]{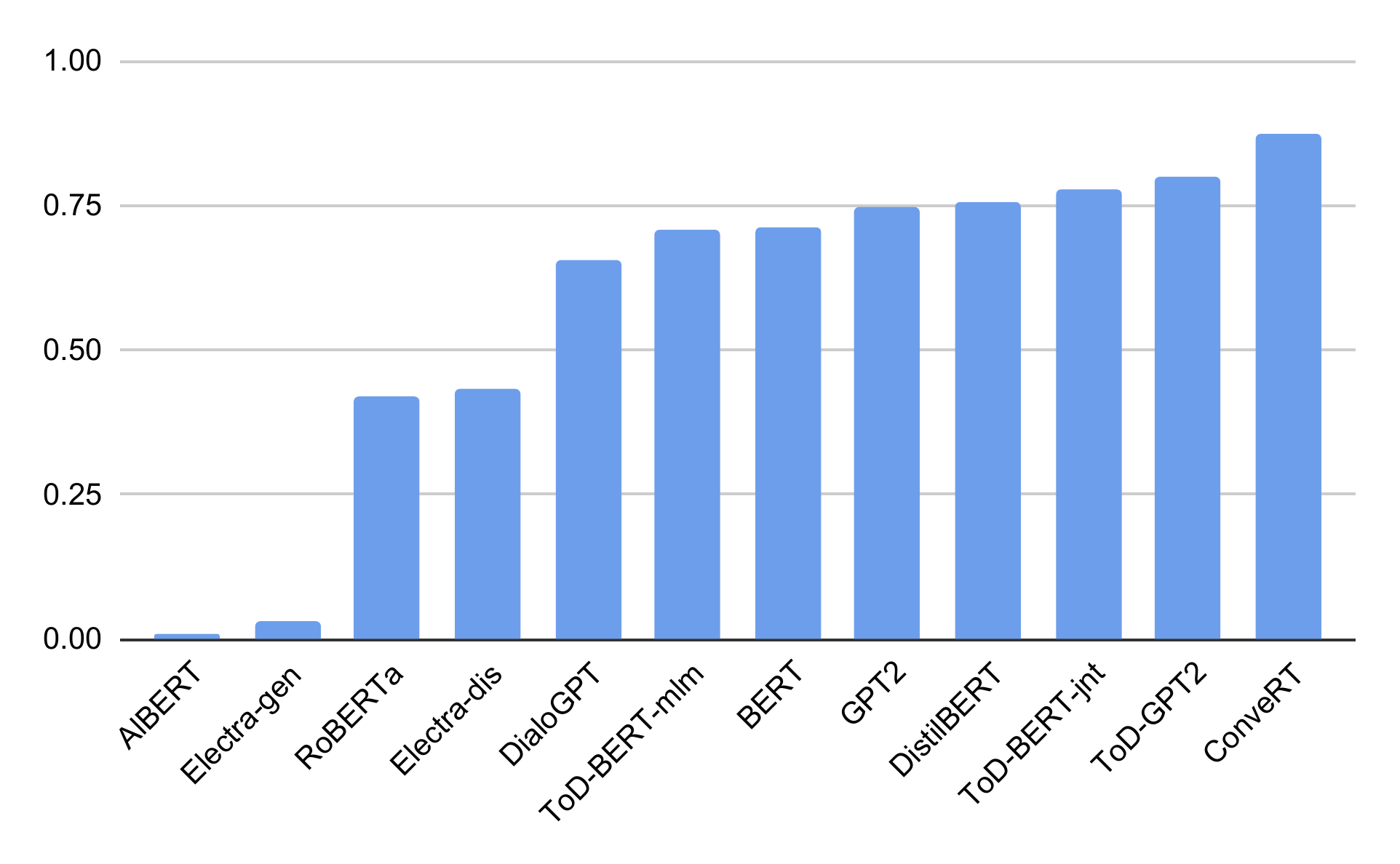}} \\
    
    \subfloat[MWOZ Slot (System)]{
        \includegraphics[width=0.9\linewidth]{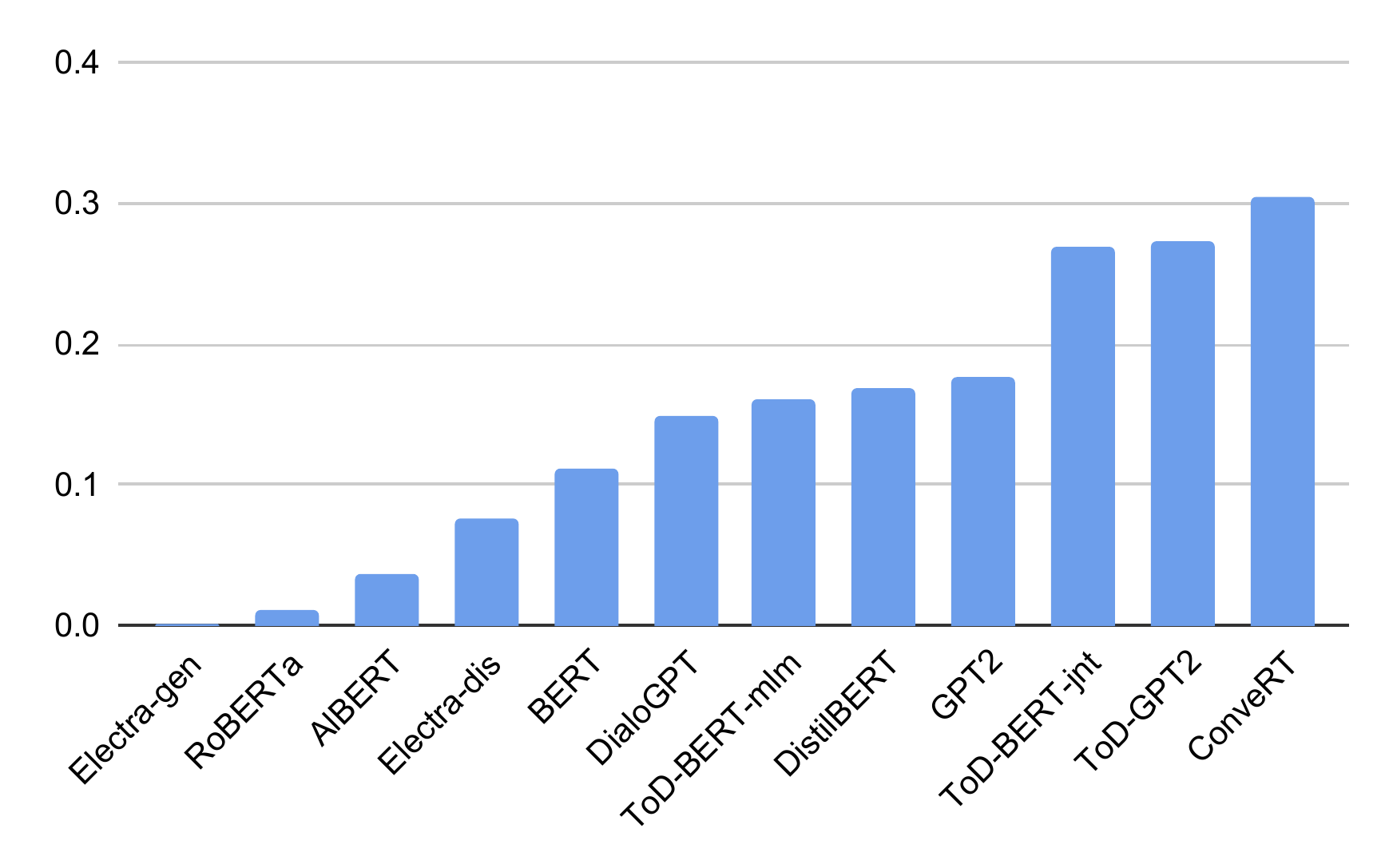}} \\
    \subfloat[MWOZ Act (System)]{
        \includegraphics[width=0.9\linewidth]{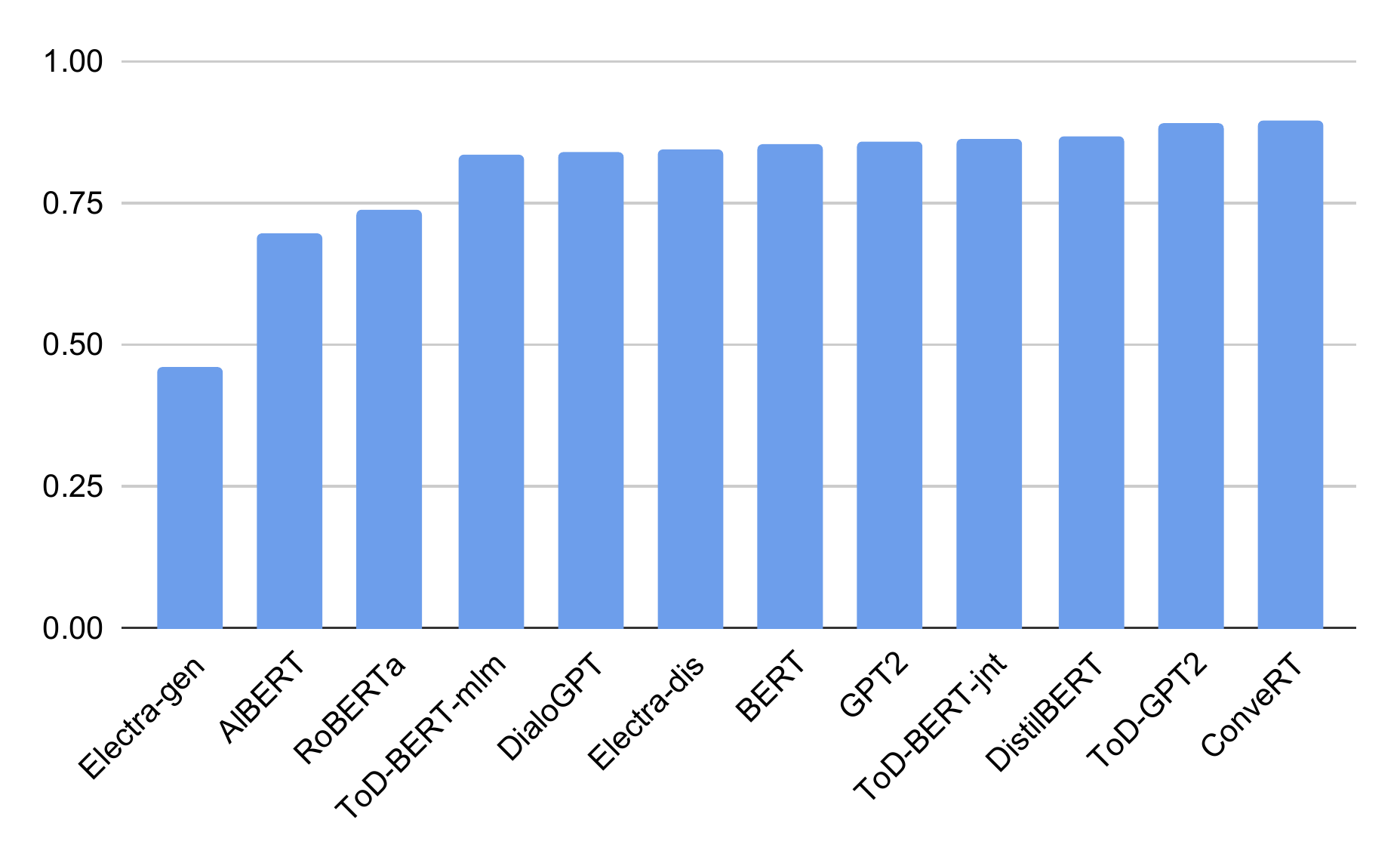}}
        
    \caption{The results of supervised classifier probe. The y-axis in (a) and (b) represents the accuracy. The y-axis in (c) and (d) represents the micro-F1 score.}
    \label{fig:probe}
\end{figure}

\begin{figure*}[h!]
    \centering
    \subfloat[MWOZ Domain (User)]{
        \includegraphics[width=0.315\linewidth]{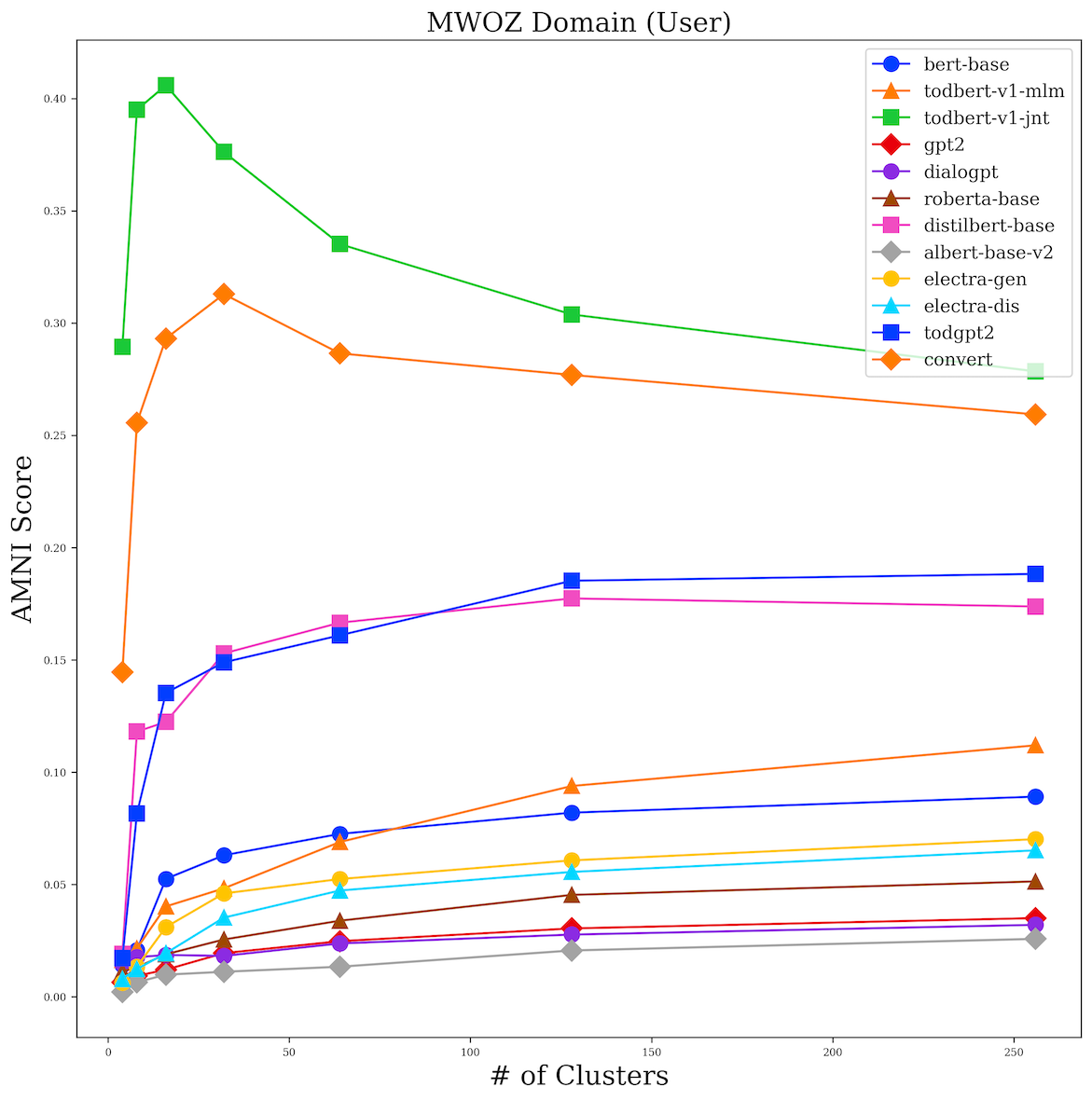}}  
    \subfloat[MWOZ Domain (System)]{
        \includegraphics[width=0.315\linewidth]{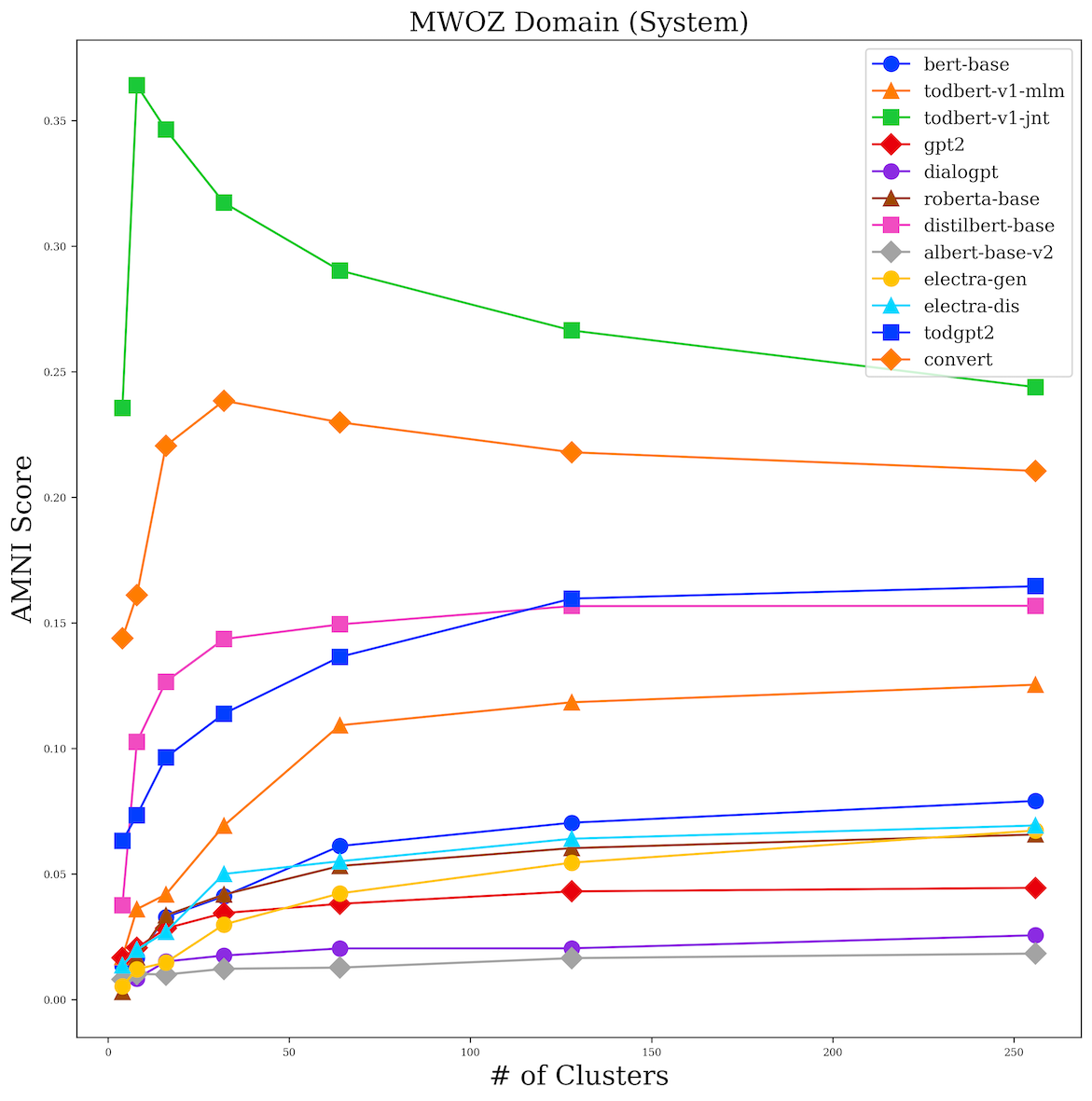}}
    \subfloat[OOS Intent (User)]{
        \includegraphics[width=0.315\linewidth]{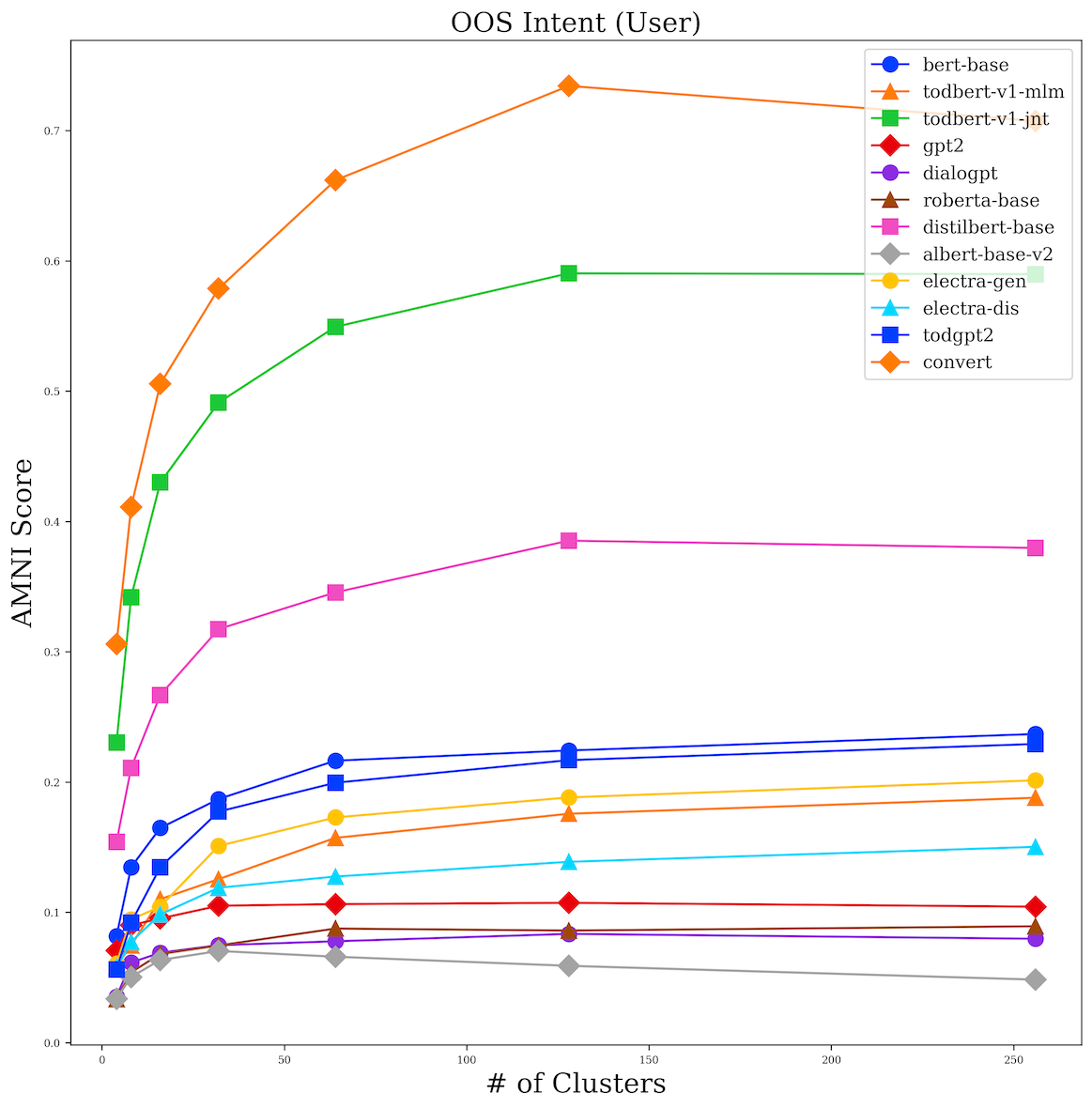}}
    \\
    
    \subfloat[MWOZ Slot - Set (User)]{
        \includegraphics[width=0.315\linewidth]{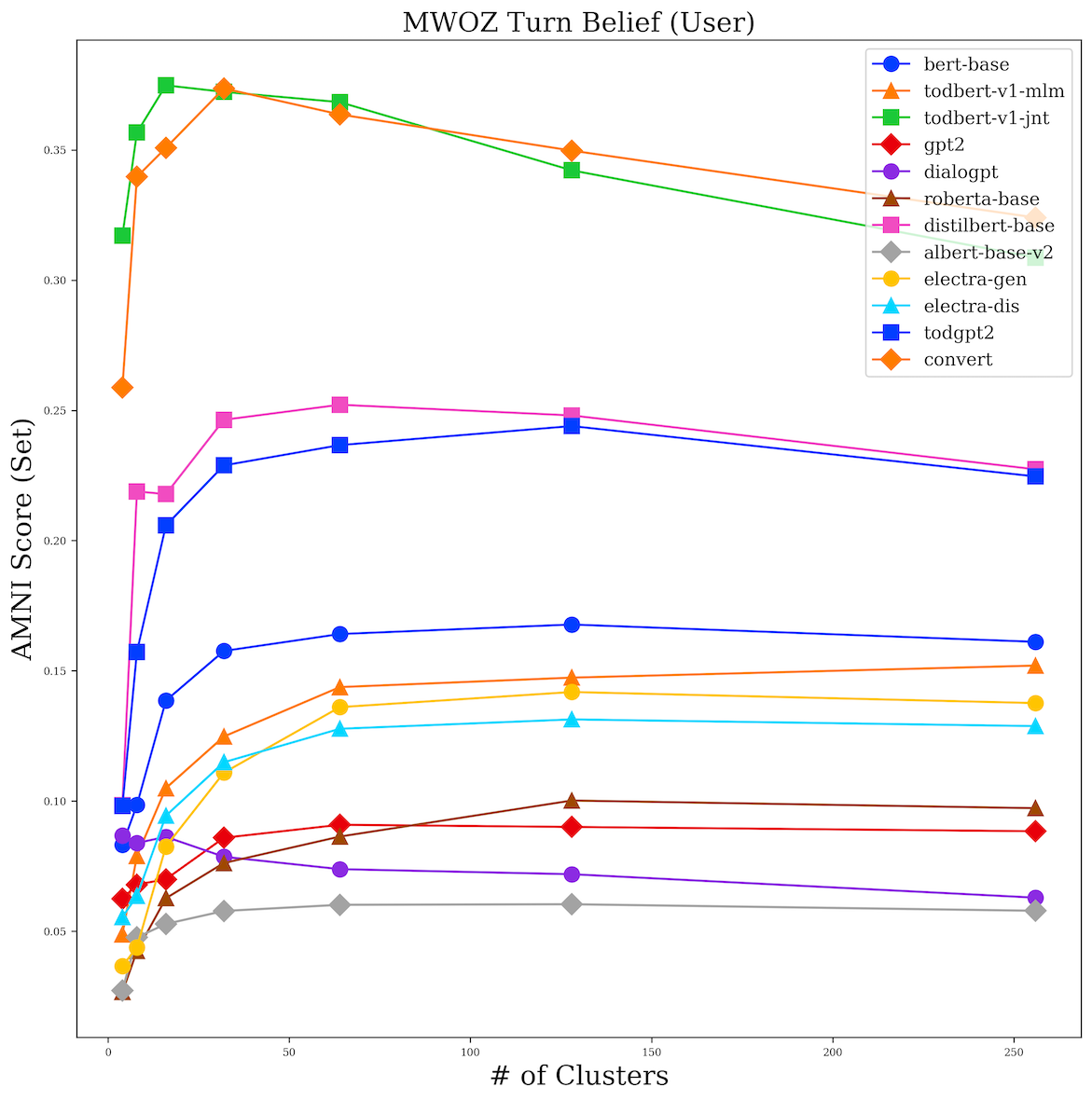}} 
    \subfloat[MWOZ Slot - Set (System)]{
        \includegraphics[width=0.315\linewidth]{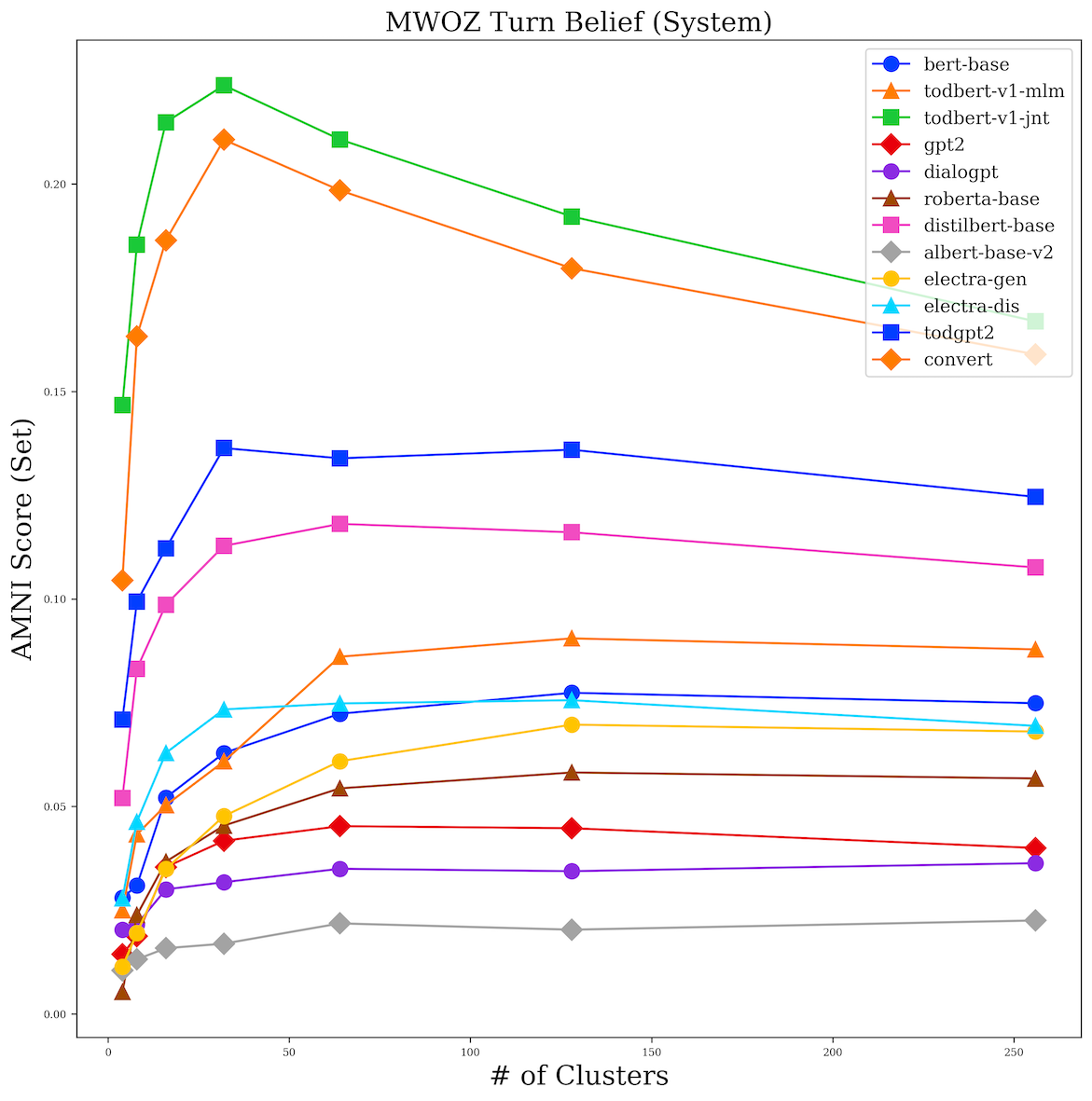}} 
    \subfloat[MWOZ Act - Set (System)]{
        \includegraphics[width=0.315\linewidth]{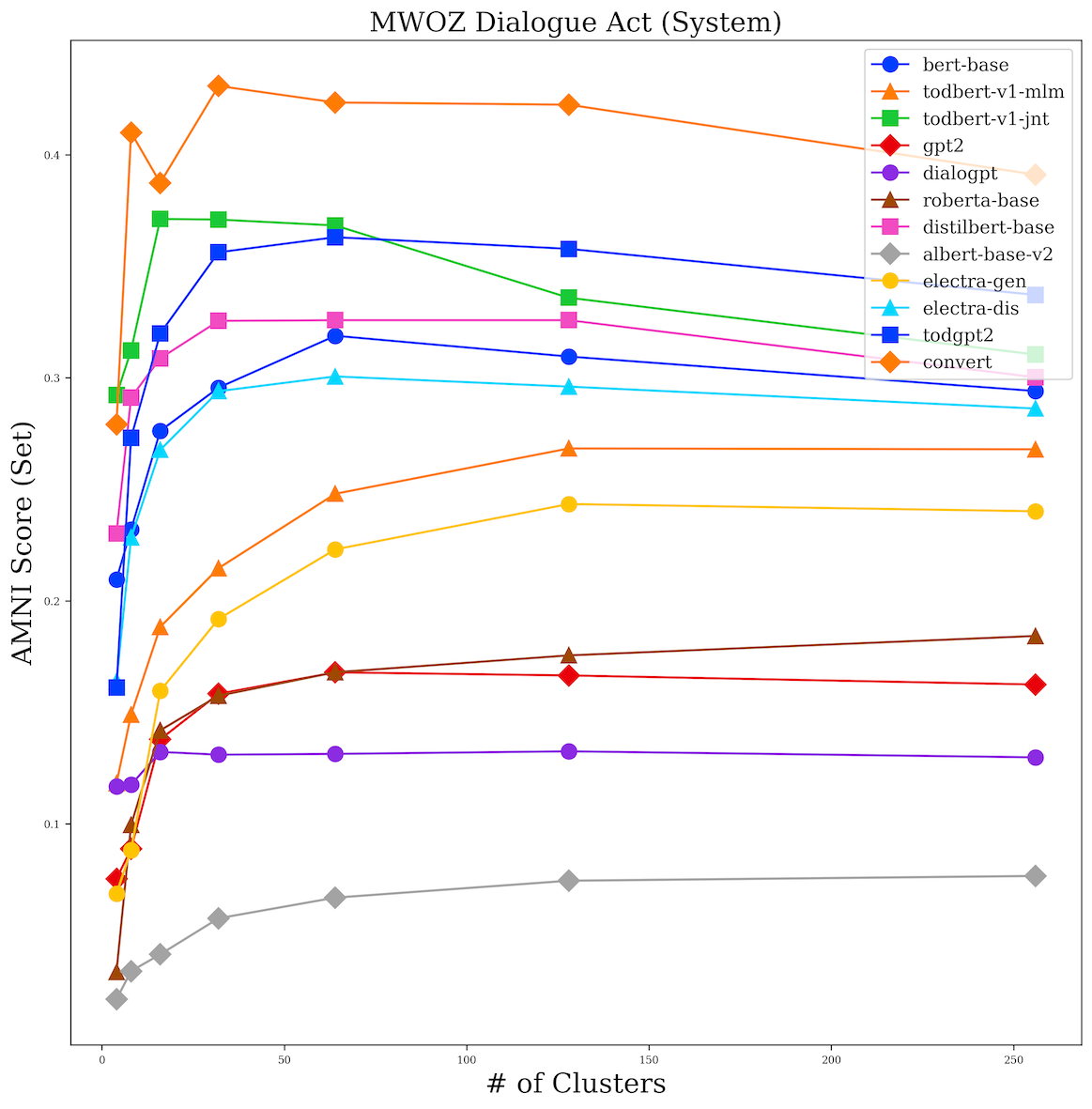}}
    \\
    
        
    \caption{The ANMI evaluation of pre-trained models with the domain, intent, slot, and action labels. The X-axis is the number of clusters and the y-axis is the ANMI score (Best view in color)}
    \label{fig:amni}
\end{figure*}

\subsection{Evaluation}
Domain identification and intent detection tasks are multi-class classification problems. Therefore, we can directly use their annotated domain and intent labels to compute the ANMI scores. Slot tagging and dialogue act prediction tasks, meanwhile, are multi-label classification problems. For example, each utterance can include multiple slots mentioned ($<$food$>$ and $<$price$>$ slots) and various actions triggered ($<$greeting$>$ and $<$inform$>$ acts). In our experiment, we use a naive way that is viewing a different set of slot or act combination as different labels, e.g., three slot sets $<$food$>$, $<$food, price$>$, and $<$price, location$>$ belong to three different clusters.

\subsection{Results}
\label{subsec-results}
\paragraph{Classifier} results are shown in Figure~\ref{fig:probe}. We can observe that ConveRT, TOD-BERT-jnt, and TOD-GPT2 achieve the best performance, implying that pre-training with dialogue-related data captures better representations, at least in these sub-tasks.
Moreover, the performance of ConveRT and TOD-BERT-jnt suggests that it is helpful to pre-train with a response selection contrastive objective, especially when comparing TOD-BERT-jnt to TOD-BERT-mlm.
Moreover, most of the pre-trained models have a similar and high micro-F1 score in (d) system dialogue act prediction, as most of them are above 75\% over 13 classes. 
Dialogue slot (c) information, meanwhile, is not well captured by these representations, resulting in a micro-F1 lower than 30\%.
On the other hand, ELECTRA-GEN, RoBERTa, and AlBERT show the worst classification results. Especially in (b) intent classification and (c) dialogue slot tagging, some of them seem to have zero useful information to make a prediction.

\paragraph{Mutual information} results using K-means clustering are shown in Figure~\ref{fig:amni}. Due to the space limit, we report the results using GMM in the Appendix, as the two of them have similar trends. The x-axis is the number of clusters in each subplot, ranging from 4 to 256, and the y-axis is the ANMI score between a predicted clustering and its corresponding true clustering. In general, the mutual information probe results are similar to what we observe in the classifier probe. We can find that ToD-BERT-jnt and ConveRT are those with the highest mutual information, and they are usually followed by TOD-GPT2 and DistilBERT. 

Another observation is that representations from those pre-trained language models, especially the top ones, seem to have more connection with user intent and system dialogue act labels than domain and slot labels. The average ANMI scores across 12 models and 7 different number of clusters for intent and dialogue act are $0.193 \pm 0.169$ and $0.226 \pm 0.107$, respectively. But domain and dialogue slot only have $0.086 \pm 0.087$ and $0.077 \pm 0.057$ AMNI scores in average. We discuss each subplot in detail in the following:

\begin{figure*}[!h]
\centering
\includegraphics[width=\linewidth]{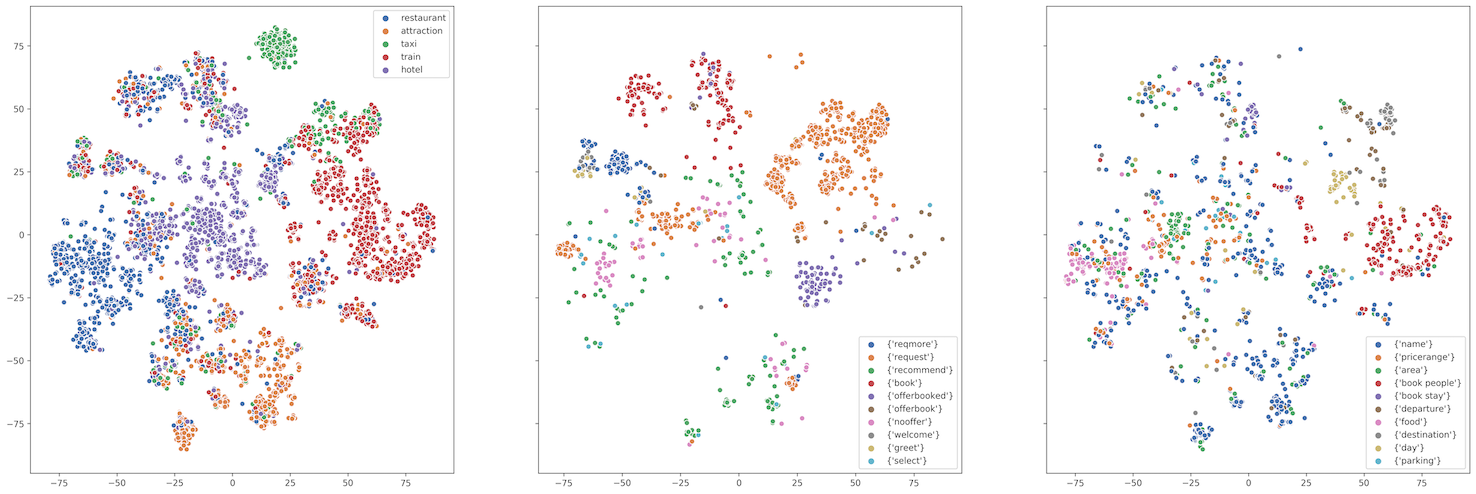}
\caption{The tSNE visualization of dialogue representations from ToD-BERT-jnt. (Best view in color)}
\label{fig:tsne-todbertjnt}
\end{figure*}
\begin{figure*}[!h]
\centering
\includegraphics[width=\linewidth]{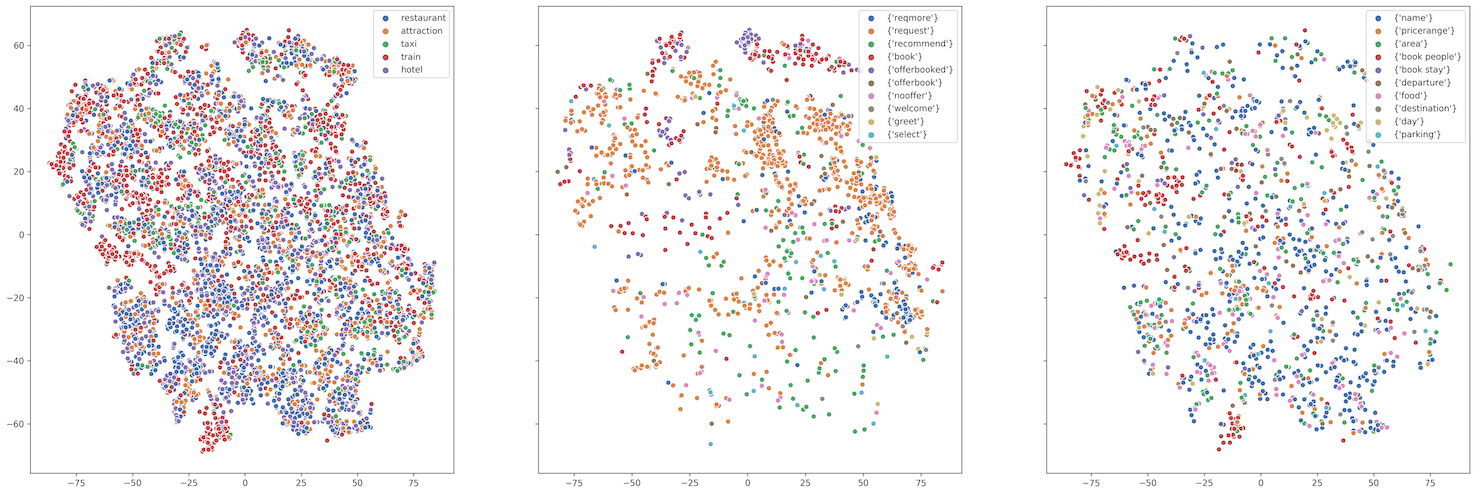}
\caption{The tSNE visualization of dialogue representations from GPT2. (Best view in color)}
\label{fig:tsne-gpt2}
\end{figure*}

Figure~\ref{fig:amni} (a) and (b) show the mutual information between predicted clustering and the true domain labels on the MWOZ dataset. A user utterance seems to have higher domain mutual information than a system response. TOD-BERT-jnt, in this case, outperforms others by a large margin, achieving around 0.4 ANMI with 8 clusters. Figure~\ref{fig:amni} (c) is about user intent using user utterances. ConveRT surpasses others by far in the mutual information of intent, achieving over 0.7 ANMI at 128 clusters when the true number of classes equals to 151. Other than the top three models (ConveRT, TOD-BERT-jnt, and DistilBERT), the remaining pre-trained models have ANMI scores lower than 0.2. 

Figure~\ref{fig:amni} (d) and (e) show the mutual information evaluation using the slot labels. When comparing (d) to (e), we can find that user utterances contain more slot information than system responses (Max around 0.35 and 0.25). It is not surprising because a user in task-oriented dialogue is usually the slot information provider, informing what location or which cuisine s/he prefers. ToD-BERT and ConveRT perform similar in this case, still outperform others by a big margin.

Figure~\ref{fig:amni} (f) shows the mutual information for the predicted clustering of system dialogue acts. We can find that most of the pre-trained language models have shown a relatively high ANMI score (average 0.226) and closed the gap between their performance and the top model. ConveRT works the best, in this case, followed by TOD-BERT-jnt and TOD-GPT2, in which two of them seem to have similar ANMI scores. 

\begin{table*}
\centering
\resizebox{0.95\linewidth}{!}{
\begin{tabular}{|c|l|}
\hline
\multicolumn{2}{|c|}{\textbf{ConveRT}} \\ \hline
\multirow{5}{*}{\begin{tabular}[c]{@{}c@{}}Cluster 1\\ (Failed Booking) \end{tabular}} & i am sorry but dojo noodle bar is solidly booked at that time . i can try a different time or day for you . \\ \cline{2-2} 
 & 1 moment while i try to make the reservation of table for 8 , friday at 16:30 . \\ \cline{2-2} 
 & booking was unsuccessful . can you try another time slot ? \\ \cline{2-2} 
 & i am very sorry i was unable to book at acorn guest house for 5 nights , would you like to try for a shorter stay ? \\ \cline{2-2} 
 & i am afraid that booking is unsuccessful . would you like a different day or amount of days ? \\ \hline
\multirow{5}{*}{\begin{tabular}[c]{@{}c@{}}Cluster 2\\ (Train Time)\end{tabular}} & there are 5 trains available , may i book 1 for you that leaves at 7:40 and arrives at 10:23 ? \\ \cline{2-2} 
 & tr0330 departs at 14:09 and arrives by 15:54 . would you like a ticket ? \\ \cline{2-2} 
 & the tr2141 arrives by 15:27 . would you like me to reserve some seats for you ? \\ \cline{2-2} 
 & i have train tr4283 that leaves cambridge at 5:29 and arrives in bishops stortford at 6:07 . would you like to make reservations ? \\ \cline{2-2} 
 & i have a train that leaves cambridge 14:01 arriving in birmingham new street at 16:44 . would that work ? \\ \hline
\multirow{5}{*}{\begin{tabular}[c]{@{}c@{}}Cluster 3\\ (Restaurant Request)\end{tabular}} & there are 21 restaurant -s available in the centre of town . how about a specific type of cuisine ? \\ \cline{2-2} 
 & there are 9 indian restaurant -s in centre what price range do you want ? \\ \cline{2-2} 
 & i am sorry , there are no catalan dining establishments in the city centre . would you like to look for a different cuisine or area ? \\ \cline{2-2} 
 & i found 4 restaurant -s with the name tandoori that serve indian food on the south , west , and east . do you have a location preference ? \\ \cline{2-2} 
 & there are no singaporean restaurant -s , but there are cheap ones offering several different cuisines . \\ \hline
 
\multirow{5}{*}{\begin{tabular}[c]{@{}c@{}}Cluster 4\\ (Confirm Booking)\end{tabular}} 
& all set . your reference number is k2bo09vq . \\ \cline{2-2} 
& i have got you booked for 16:30 . the reference number is eq0yaq1g . \\ \cline{2-2} 
& your reservation was a success and the reference number is jtwxfm7m . \\ \cline{2-2} 
& i have got your booking set , the reference number is 9rmfgjma . \\ \cline{2-2} 
& i booked tr3932 , reference number is fiw5abo2 . \\ \hline

\multirow{5}{*}{\begin{tabular}[c]{@{}c@{}}Cluster 5\\ (Hotel Request)\end{tabular}} 
& what part of town there are none in the west . \\ \cline{2-2} 
&  i can help you with that . do you have any special area you would like to stay ? or possibly a star request ? \\ \cline{2-2} 
& there are no colleges close to the area you are requesting , would you like to chose another destination ? \\ \cline{2-2} 
& sure , what area are you thinking of staying ? \\ \cline{2-2} 
& i would be happy to help . may i ask what price range and area of town you are looking for ? \\ \hline
 
\end{tabular}
}
\caption{Clustering results of the ConveRT model. The samples are picked from each randomly selected five clusters with K=32. We can roughly label a topic for each cluster.}
\label{tab:convert}
\end{table*}

\section{More Analysis}

\paragraph{Difference Between Probes}
Ideally, both probes should distinguish the goodness of different pre-trained language models, i.e., features that can be easily classified or features with high correlation with true distributions are preferred.
However, we found that although, in general, the trends we observe from two probing methods are similar, they are not the same in terms of the ranking.
When comparing the ranking of GPT2 and DialoGPT models in Figure~\ref{fig:probe} and Figure~\ref{fig:amni}, we found that they obtain almost the worse ANMI scores but work quite good in classification accuracy. This observation means that their representations of different classes are ``close'' to each other as a low ANMI score suggesting a more noisy clustering. Still, at the same time, it is not hard to find a hyperplane that can well discriminate those features.

\begin{figure}[t]
\centering
\includegraphics[width=0.85\linewidth]{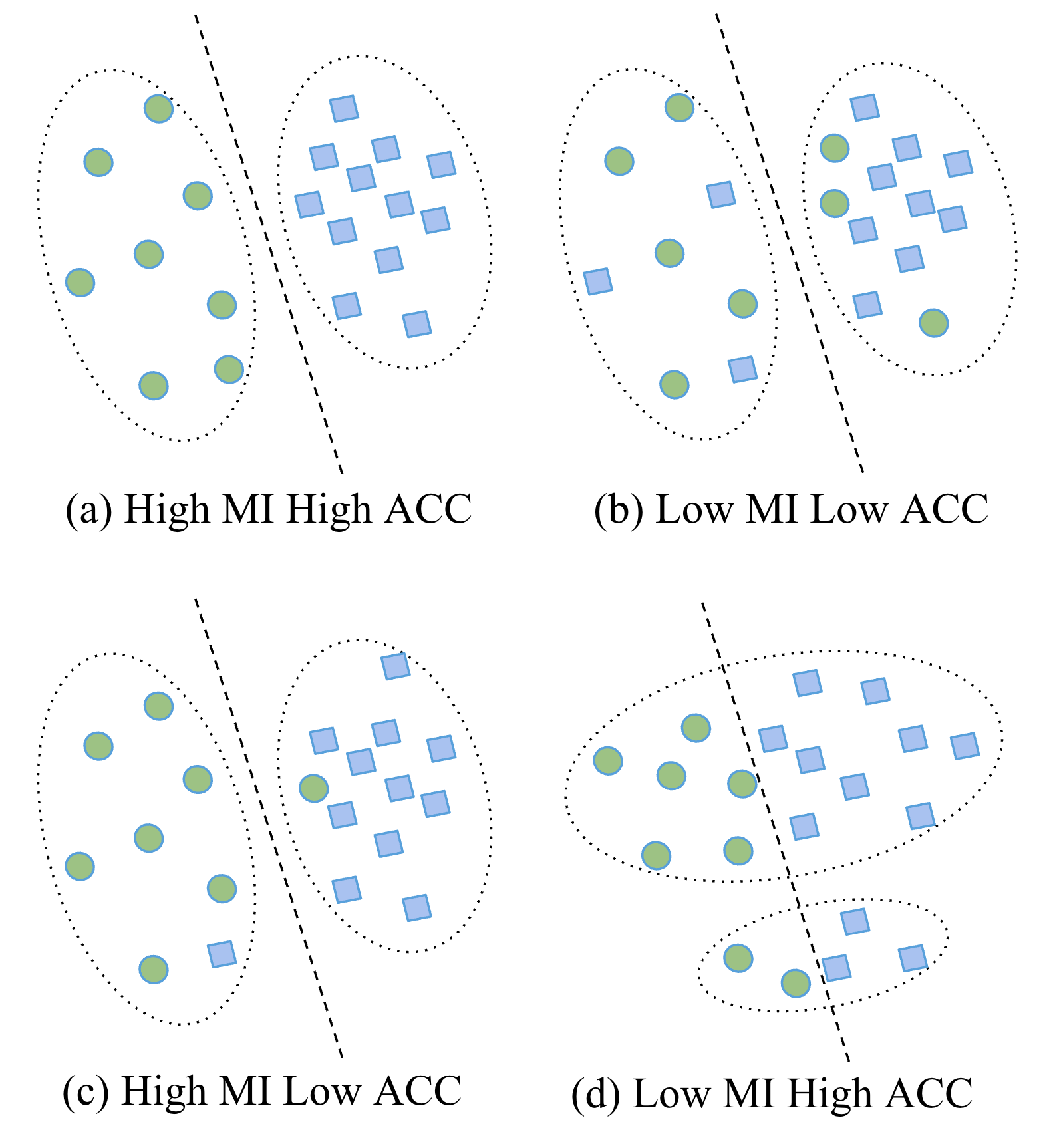}
\caption{Illustration of four different type of clusterings related to mutual information and accuracy.}
\label{fig:intuition}
\end{figure}

We discuss some possible reasons for this interesting observation in the following.
The first guess is that these features may not follow a Gaussian distribution, as we assume during clustering, suggesting that more advanced clustering techniques can be investigated in future work.
The second guess is that these features have an unavoidable clustering noise that can be denoised or debiased easily by a strong supervision signal.
The third guess, which may be a possible reason, is that these features are clustered by some other factors that are not tested, and at the same time, the factors we are interested in are scattered in groups for different classes in a similar way. Intuitively, there are four clustering results shown in Figure~\ref{fig:intuition}, where GPT2 and DialoGPT may fall into the (d) clustering type, which has a lower mutual information score but higher classification accuracy.

As a result, we suggest a simple rule of thumb regarding which probing results. In short, the results of the classifier probe could be useful if a supervised approach for a downstream task is designed, e.g., user dialogue act prediction and dialogue state tracking. On the other hand, the mutual information probe is more effective for an unsupervised problem, e.g., utterance clustering and dialogue parsing tasks.

\paragraph{Visualization}
In Figure~\ref{fig:tsne-todbertjnt} and Figure~\ref{fig:tsne-gpt2}, we visualize the embeddings of TOD-BERT-jnt and GPT2 given the same system responses from the MWOZ test set. Each point is reduced from its high-dimension features to a two-dimension point using the t-distributed stochastic neighbor embedding (tSNE). We use different colors to represent different domains (left), dialogue acts (middle), and turn slots (right). As one can observe, TOD-BERT-jnt has more clear group boundaries and better clustering results than GPT2. Visualization plots for other pre-trained models are shown in the Appendix.

\paragraph{What utterances are clustered together?}
In Table~\ref{tab:convert}, we show the clustering examples of system responses from the top performance model ConveRT. We use $K=32$ clustering and randomly select five clusters and five samples each. We found that most of the utterances are related to an unsuccessful booking in the cluster 1, containing ``I am sorry,'' ``solidly booked,'' or ``booking was unsuccessful.'' We also found other clusters showing good clustering results, such as selecting departure or arrival time for a train ticket or requesting more user preference for a restaurant reservation. More clustering results are shown in the Appendix.


\section{Conclusion}
We investigate representations from pre-trained language models for task-oriented dialogue tasks, including domain identification, intent detection, slot tagging, and dialogue act prediction. We use a supervised classifier probe and a proposed unsupervised mutual information probe. From the ranking results of two different probings, we show a list of interesting observations to provide model selection guidelines and shed light on future research towards a more advanced language modeling learning for dialogue applications.



\bibliography{emnlp2020}
\bibliographystyle{acl_natbib}

\newpage
\clearpage

\appendix

\begin{table*}
\centering
\begin{tabular}{|c|}
\hline
OOS Intent \\ \hline
\begin{tabular}[c]{@{}c@{}}'translate', 'transfer', 'timer', 'definition', 'meaning\_of\_life', 'insurance\_change', 'find\_phone', \\ 'travel\_alert', 'pto\_request', 'improve\_credit\_score', 'fun\_fact', 'change\_language', 'payday', \\ 'replacement\_card\_duration', 'time', 'application\_status', 'flight\_status', 'flip\_coin', \\ 'change\_user\_name', 'where\_are\_you\_from', 'shopping\_list\_update', 'what\_can\_i\_ask\_you', \\ 'maybe', 'oil\_change\_how', 'restaurant\_reservation', 'balance', 'confirm\_reservation', \\ 'freeze\_account', 'rollover\_401k', 'who\_made\_you', 'distance', 'user\_name', 'timezone', \\ 'next\_song', 'transactions', 'restaurant\_suggestion', 'rewards\_balance', 'pay\_bill', \\ 'spending\_history', 'pto\_request\_status', 'credit\_score', 'new\_card', 'lost\_luggage', 'repeat', \\ 'mpg', 'oil\_change\_when', 'yes', 'travel\_suggestion', 'insurance', 'todo\_list\_update', 'reminder', \\ 'change\_speed', 'tire\_pressure', 'no', 'apr', 'nutrition\_info', 'calendar', 'uber', 'calculator', 'date', \\ 'carry\_on', 'pto\_used', 'schedule\_maintenance', 'travel\_notification', 'sync\_device', 'thank\_you',\\  'roll\_dice', 'food\_last', 'cook\_time', 'reminder\_update', 'report\_lost\_card', 'ingredient\_substitution',\\  'make\_call', 'alarm', 'todo\_list', 'change\_accent', 'w2', 'bill\_due', 'calories', 'damaged\_card', \\ 'restaurant\_reviews', 'routing', 'do\_you\_have\_pets', 'schedule\_meeting', 'gas\_type', 'plug\_type',\\  'tire\_change', 'exchange\_rate', 'next\_holiday', 'change\_volume', 'who\_do\_you\_work\_for',\\  'credit\_limit', 'how\_busy', 'accept\_reservations', 'order\_status', 'pin\_change', 'goodbye', \\ 'account\_blocked', 'what\_song', 'international\_fees', 'last\_maintenance', 'meeting\_schedule',\\  'ingredients\_list', 'report\_fraud', 'measurement\_conversion', 'smart\_home', 'book\_hotel',\\  'current\_location', 'weather', 'taxes', 'min\_payment', 'whisper\_mode', 'cancel', 'international\_visa', \\ 'vaccines', 'pto\_balance', 'directions', 'spelling', 'greeting', 'reset\_settings', 'what\_is\_your\_name',\\  'direct\_deposit', 'interest\_rate', 'credit\_limit\_change', 'what\_are\_your\_hobbies', 'book\_flight',\\  'shopping\_list', 'text', 'bill\_balance', 'share\_location', 'redeem\_rewards', 'play\_music',\\  'calendar\_update', 'are\_you\_a\_bot', 'gas', 'expiration\_date', 'update\_playlist', 'cancel\_reservation',\\  'tell\_joke', 'change\_ai\_name', 'how\_old\_are\_you', 'car\_rental', 'jump\_start', 'meal\_suggestion',\\  'recipe', 'income', 'order', 'traffic', 'order\_checks', 'card\_declined', 'oos'\end{tabular} \\ \hline
\end{tabular}
\caption{OOS intent}
\label{tab:oos_intent}
\end{table*}


\begin{table*}
\centering
\begin{tabular}{|r|c|c|c|}
\hline
\textbf{Name} & \textbf{\# Dialogue} & \textbf{\# Utterance} & \textbf{Avg. Turn} \\ \hline
MetaLWOZ & 37,884 & 432,036 & 11.4 \\ \hline
Schema & 22,825 & 463,284 & 20.3  \\ \hline
Taskmaster & 13,215 & 303,066 & 22.9  \\ \hline
MWOZ & 10,420 & 71,410 & 6.9  \\ \hline
MSR-E2E & 10,087 & 74,686 & 7.4  \\ \hline
SMD & 3,031 & 15,928 & 5.3  \\ \hline
Frames & 1,369 & 19,986 & 14.6  \\ \hline
WOZ & 1,200 & 5,012 & 4.2  \\ \hline
CamRest676 & 676 & 2,744 & 4.1 \\ \hline
\end{tabular}
\caption{The data statistics is from \citet{wu2020tod}.}
\label{tb:train_dataset}
\end{table*}

\begin{figure*}[h!]
    \centering
    \resizebox{\linewidth}{!}{
    \subfloat[MWOZ Domain (User)]{
        \includegraphics[width=0.25\linewidth]{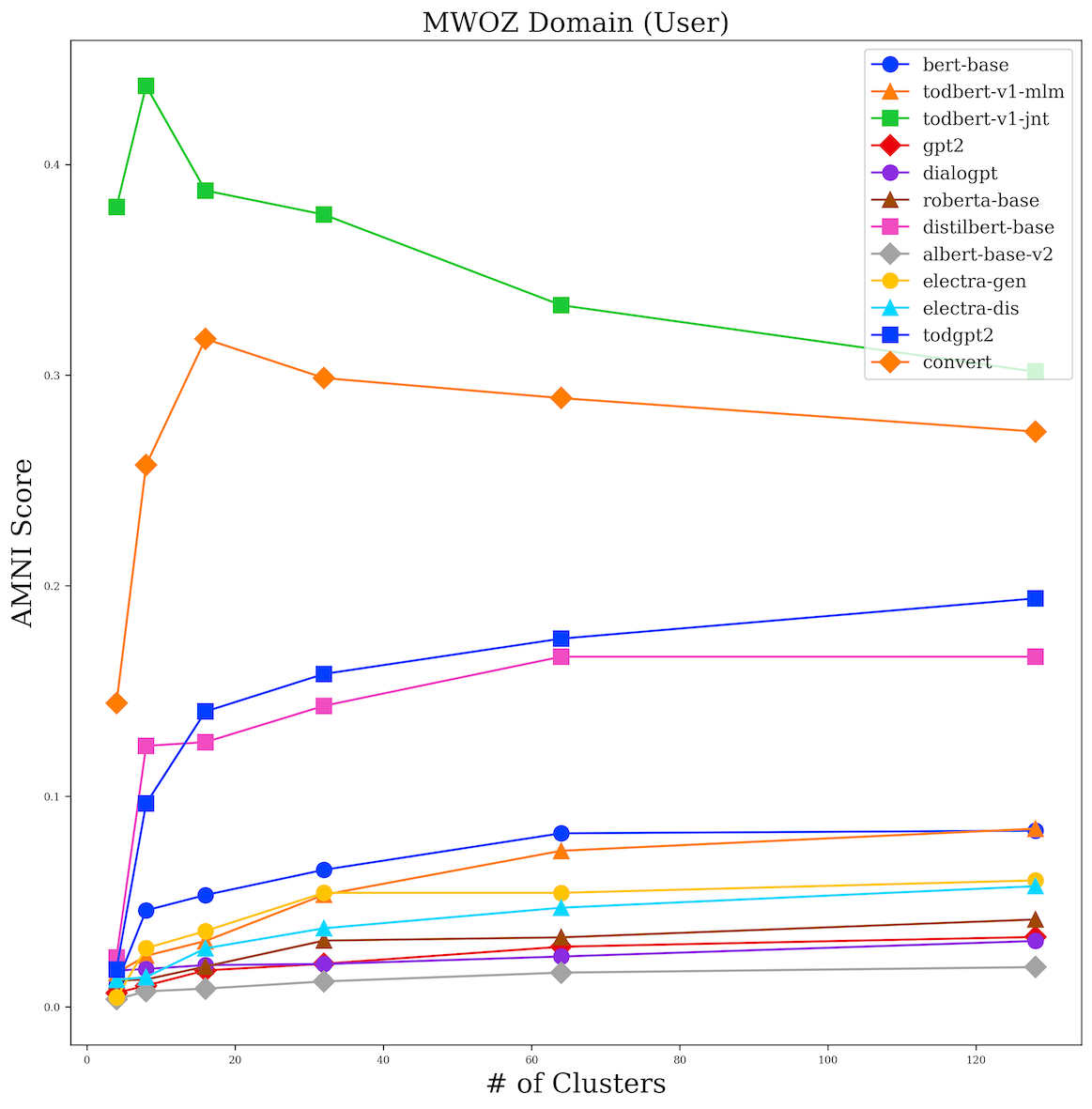}} 
    \hfill
    \subfloat[MWOZ Domain (Sys)]{
        \includegraphics[width=0.25\linewidth]{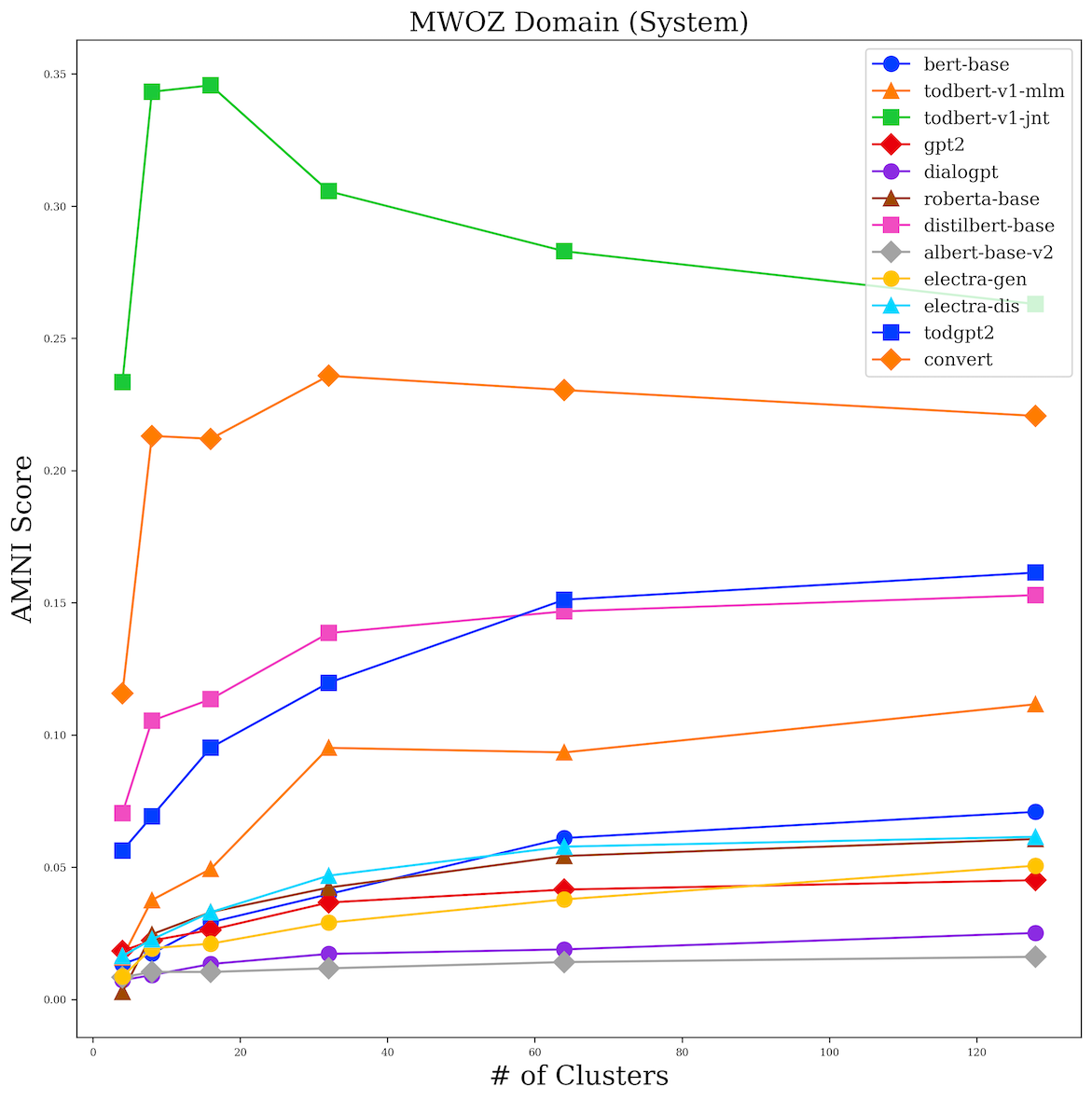}}
    \hfill
    \subfloat[OOS Intent (User)]{
        \includegraphics[width=0.25\linewidth]{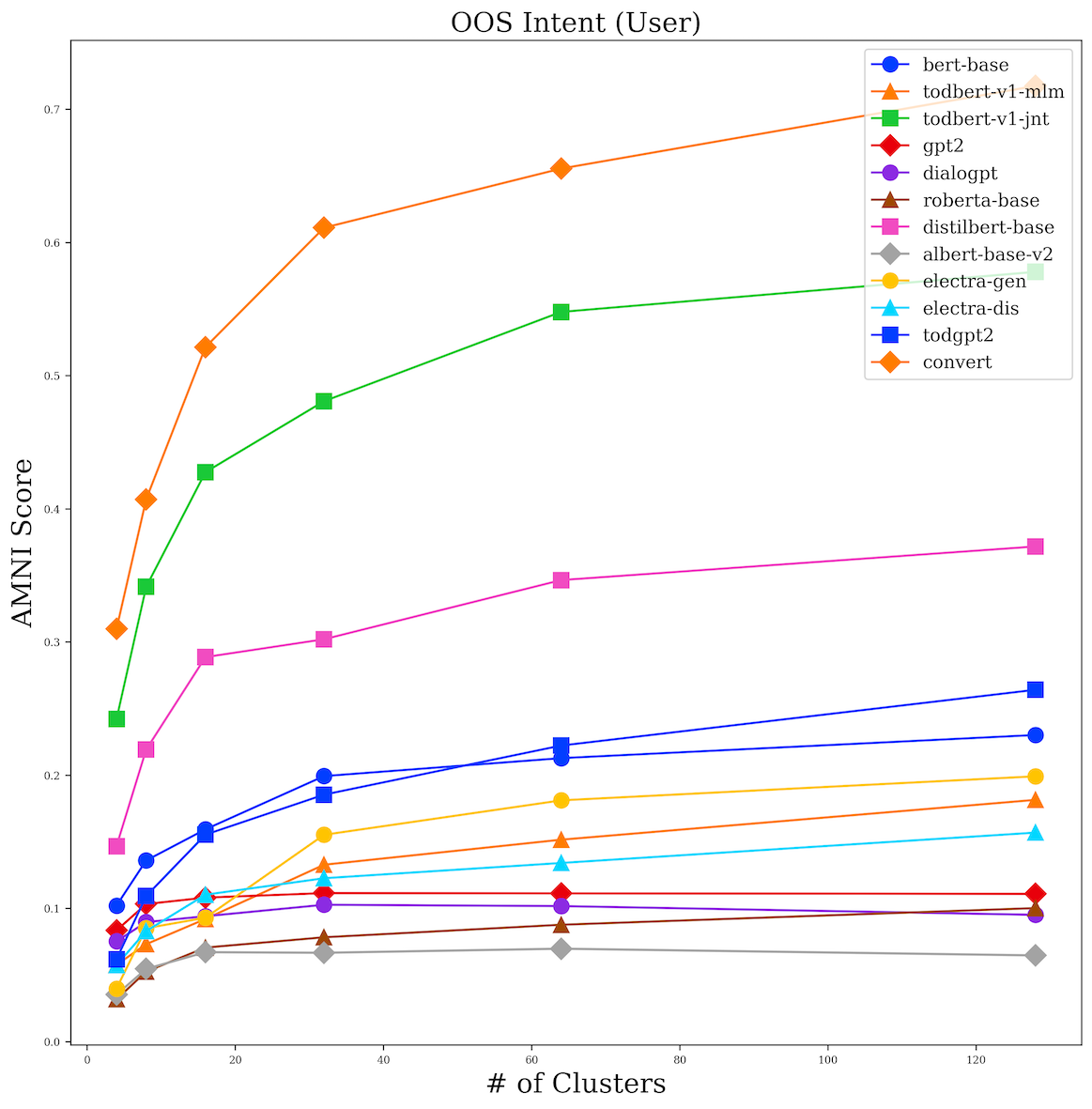}}
    }\\
    
    \resizebox{\linewidth}{!}{
    \subfloat[MWOZ Slot - Set (User)]{
        \includegraphics[width=0.25\linewidth]{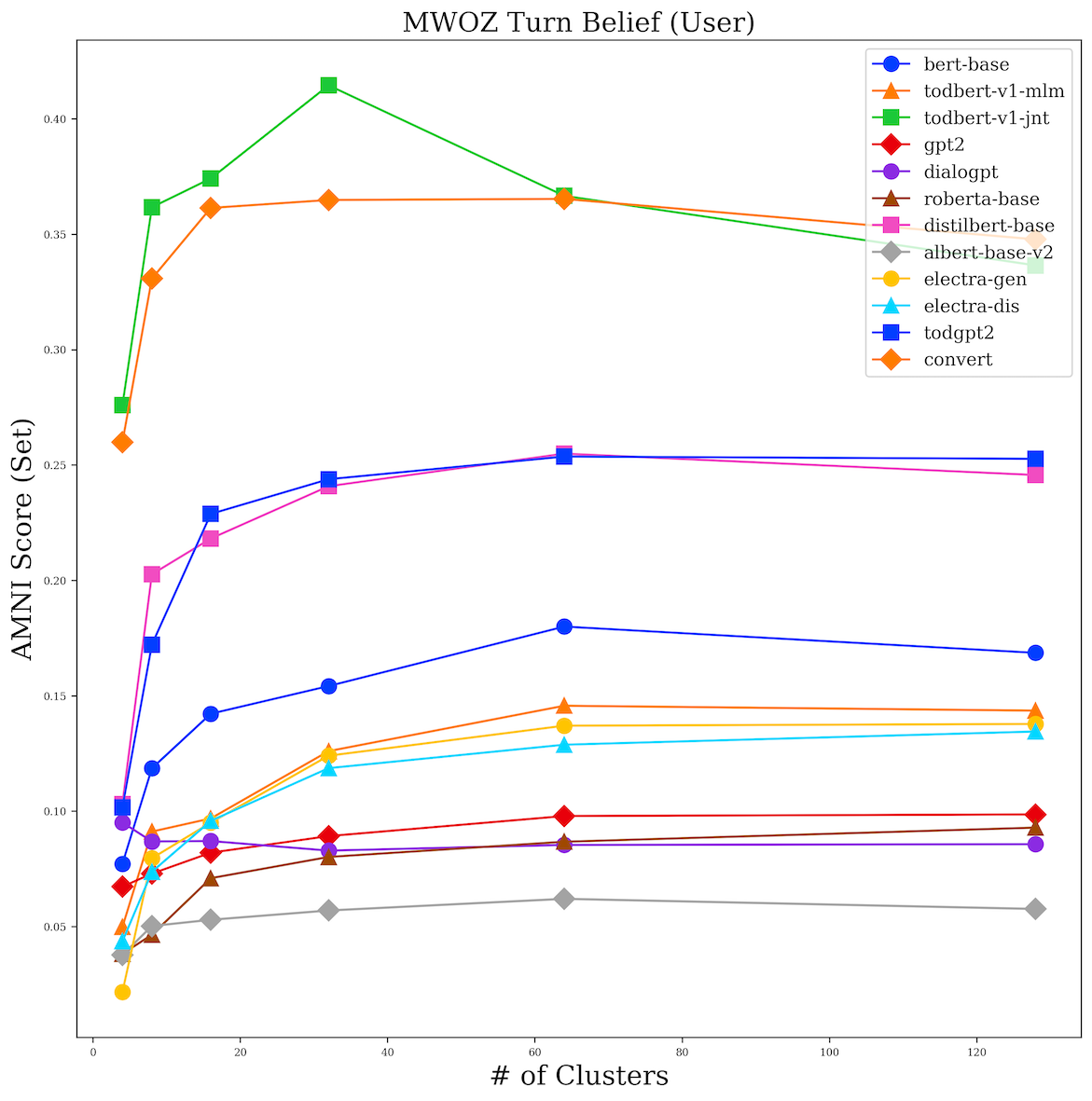}} 
    \hfill
    \subfloat[MWOZ Slot - Set (Sys)]{
        \includegraphics[width=0.25\linewidth]{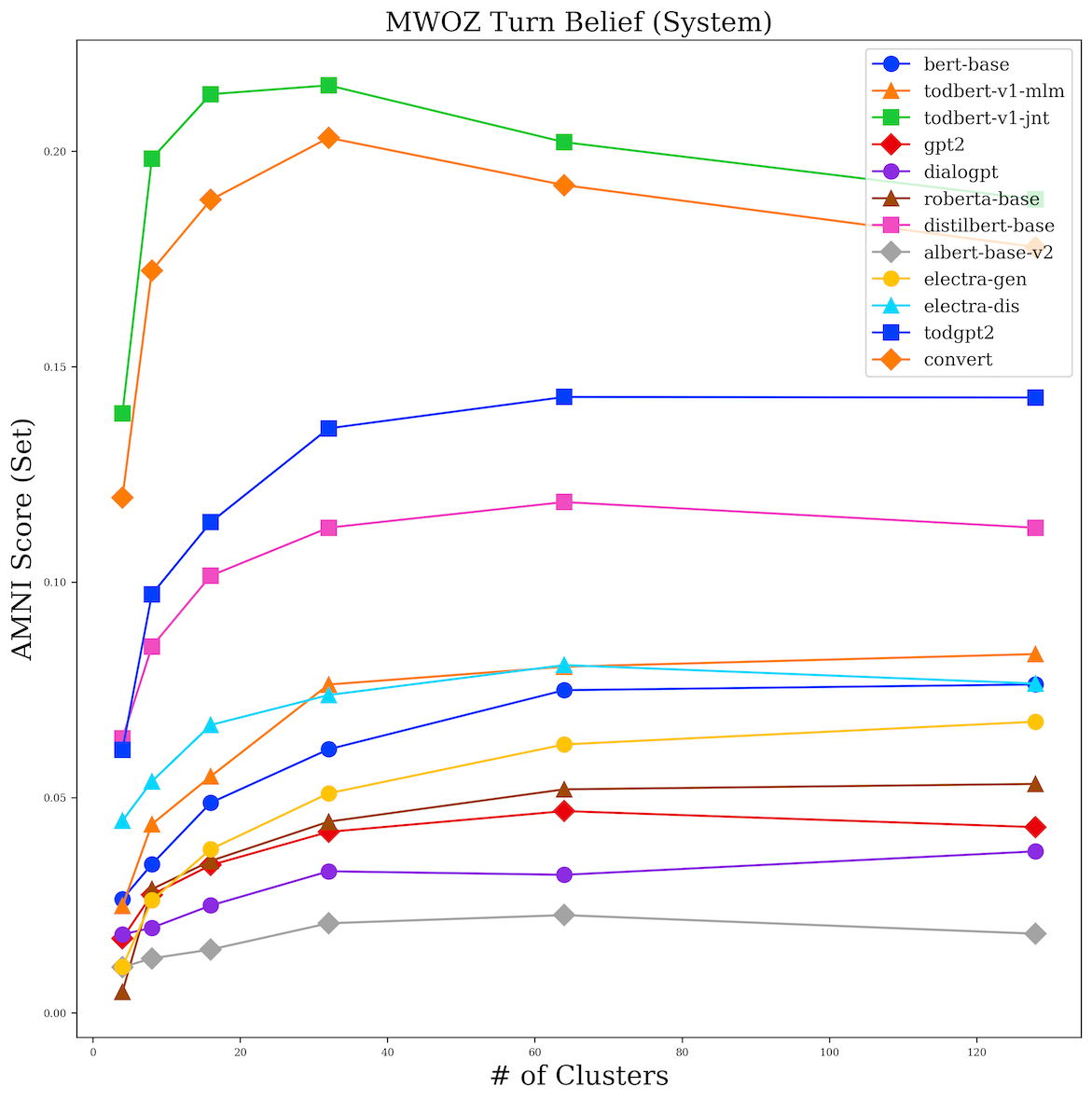}} 
    \hfill    
    \subfloat[MWOZ Act - Set (Sys)]{
        \includegraphics[width=0.25\linewidth]{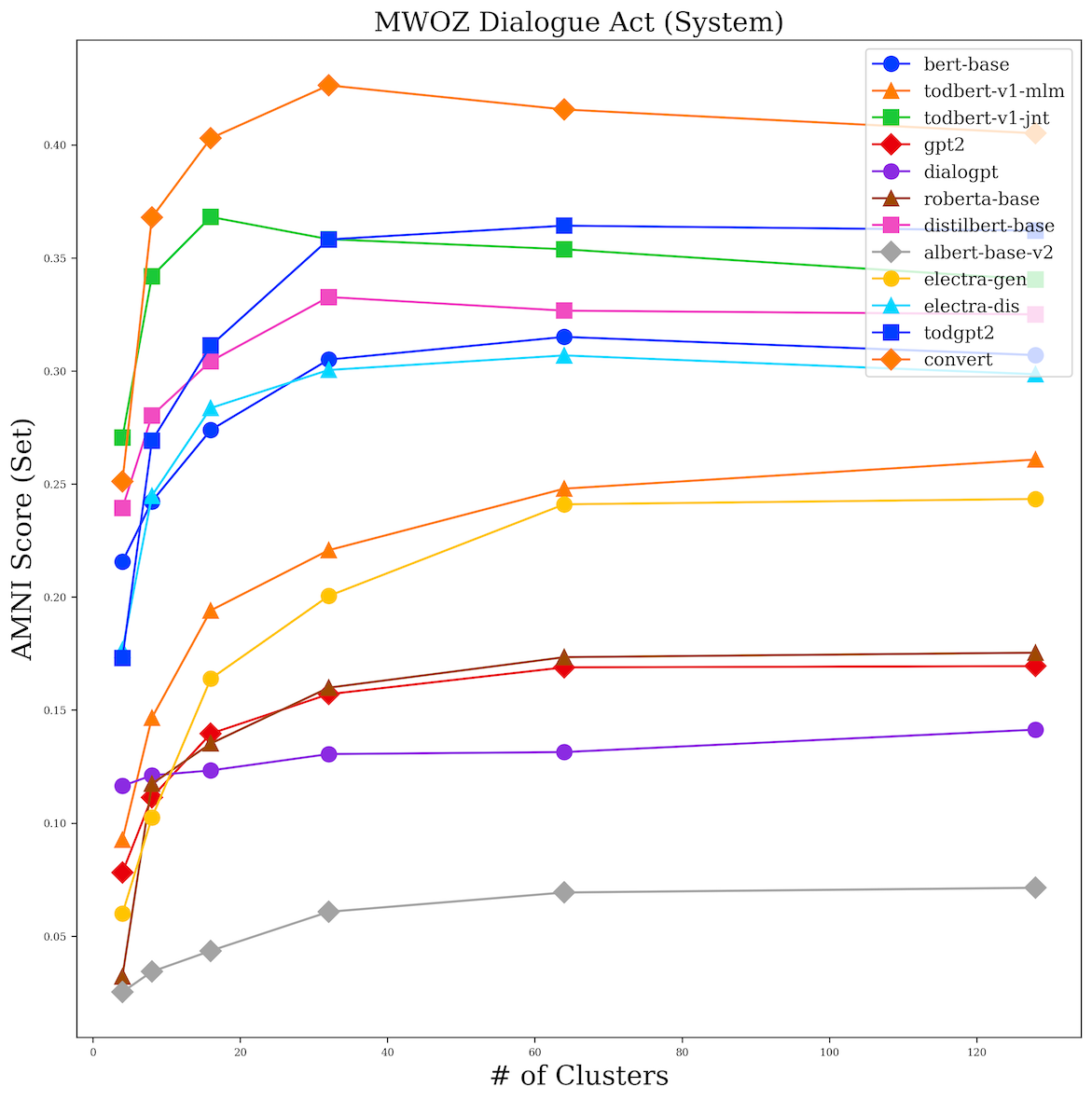}}
    }\\

    \caption{The ANMI evaluation of pre-pretrained models with domain, intent, slot, and action labels using GMM. (Best view in color)}
    \label{fig:amni-gmm}
\end{figure*}


\begin{figure*}[t]
\centering
\includegraphics[width=\linewidth]{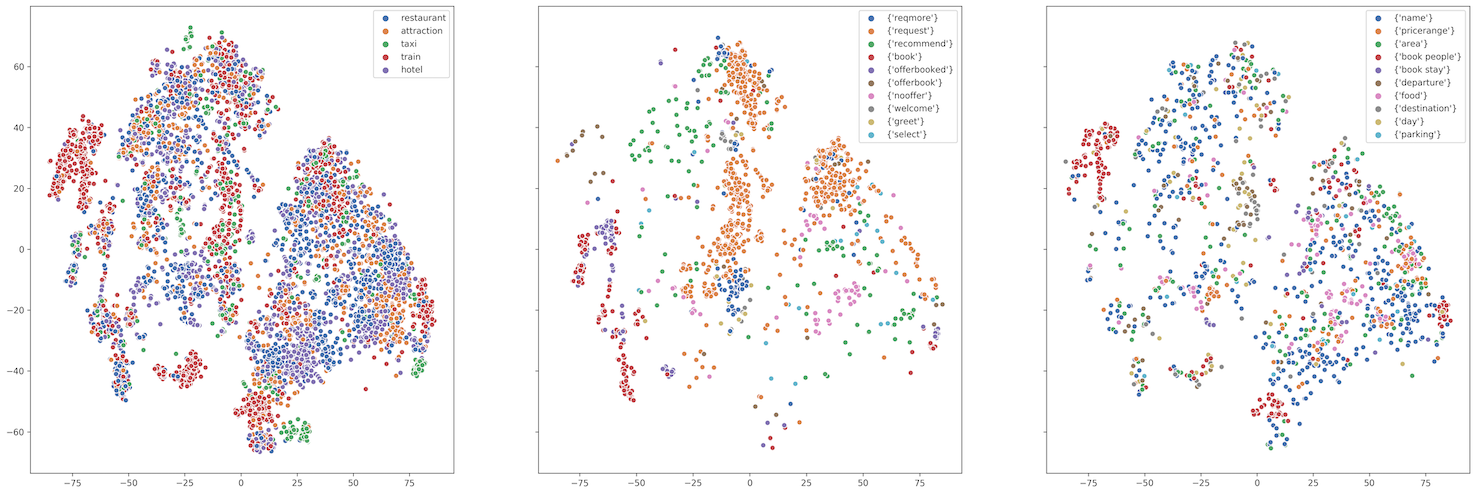}
\caption{The tSNE visualization of dialogue representations from the ToD-BERT-jnt. (Best view in color.)}
\label{fig:tsne-todbertmlm}
\end{figure*}

\begin{figure*}[t]
\centering
\includegraphics[width=\linewidth]{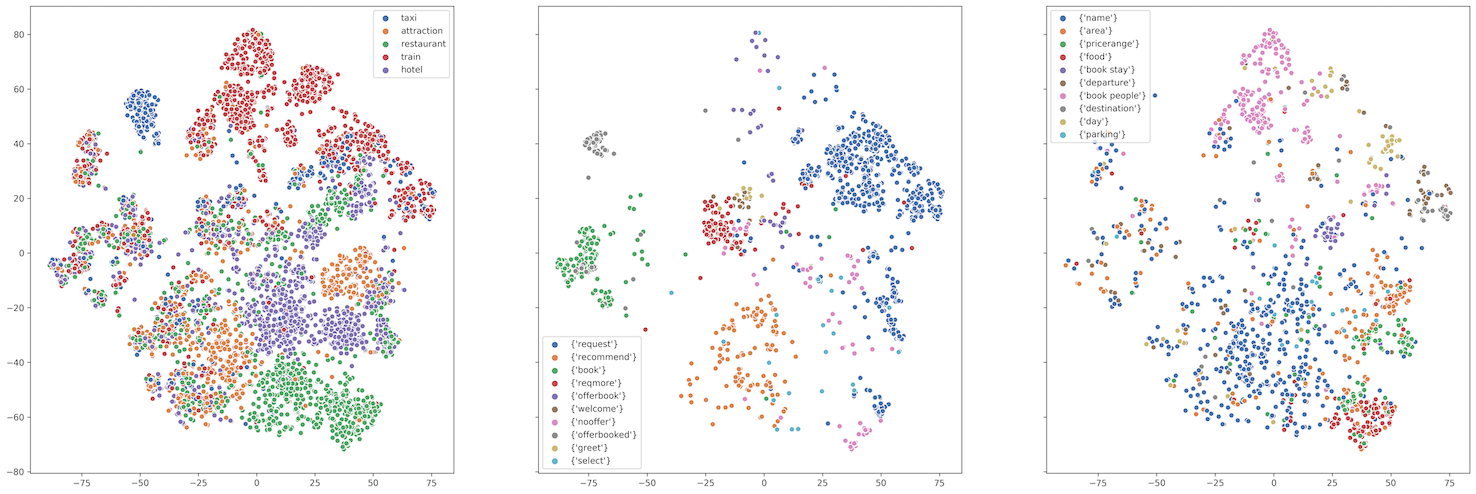}
\caption{The tSNE visualization of dialogue representations from the ConveRT. (Best view in color)}
\label{fig:tsne-convert}
\end{figure*}

\begin{figure*}[t]
\centering
\includegraphics[width=\linewidth]{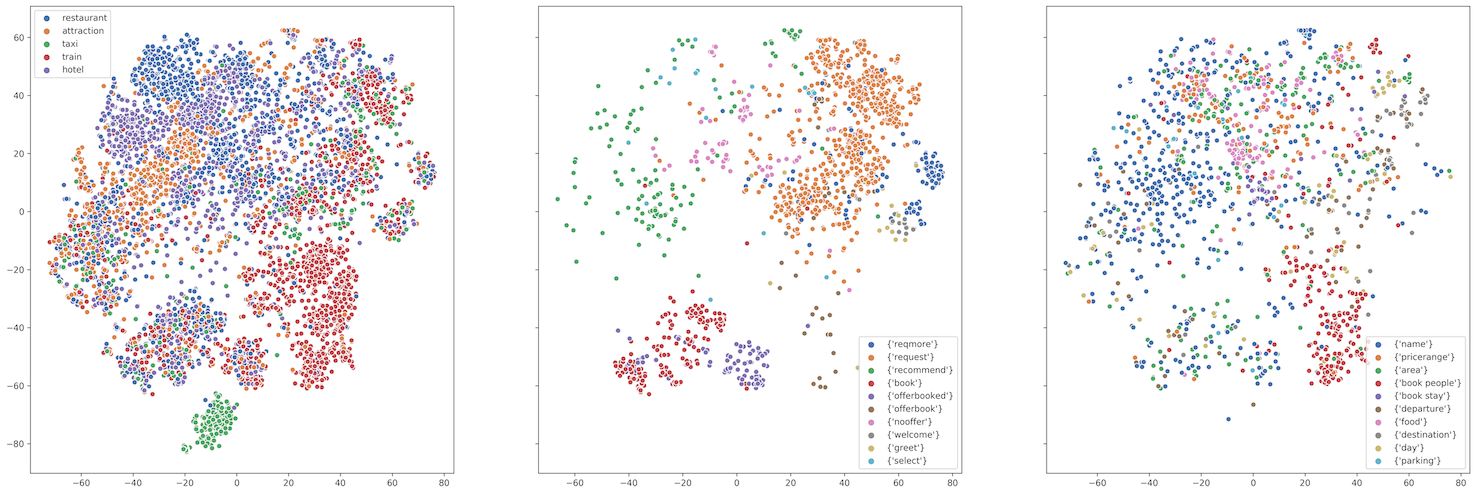}
\caption{The tSNE visualization of dialogue representations from the DistilBERT. (Best view in color.)}
\label{fig:tsne-distilbert}
\end{figure*}

\begin{figure*}[t]
\centering
\includegraphics[width=\linewidth]{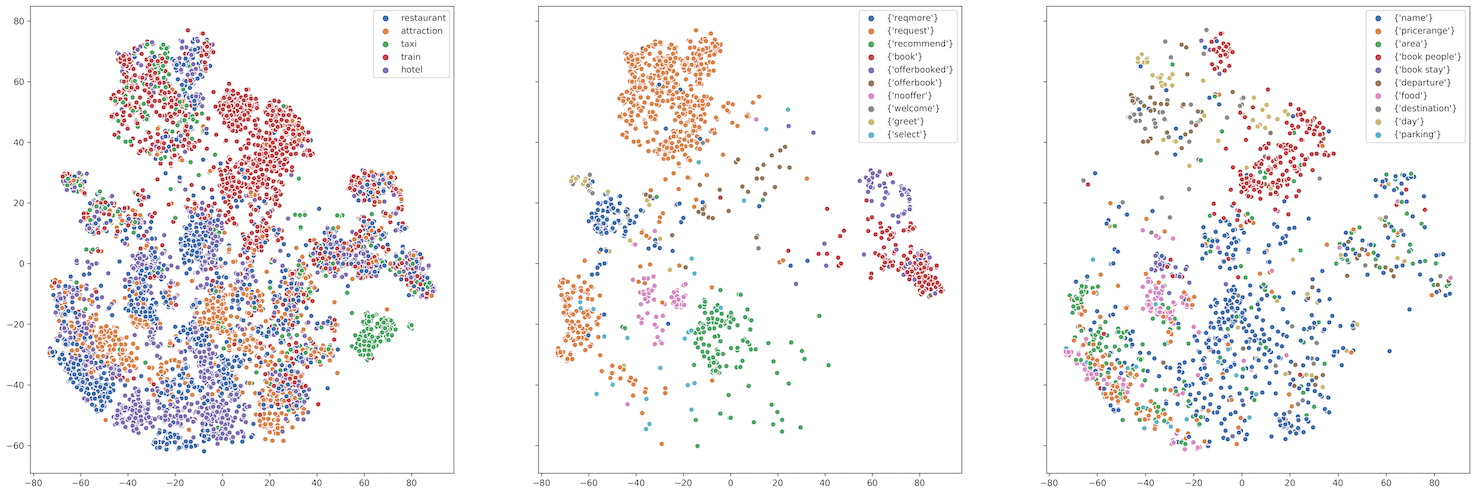}
\caption{The tSNE visualization of dialogue representations from the ToD-GPT. (Best view in color.)}
\label{fig:tsne-todgpt}
\end{figure*}

\begin{figure*}[t]
\centering
\includegraphics[width=\linewidth]{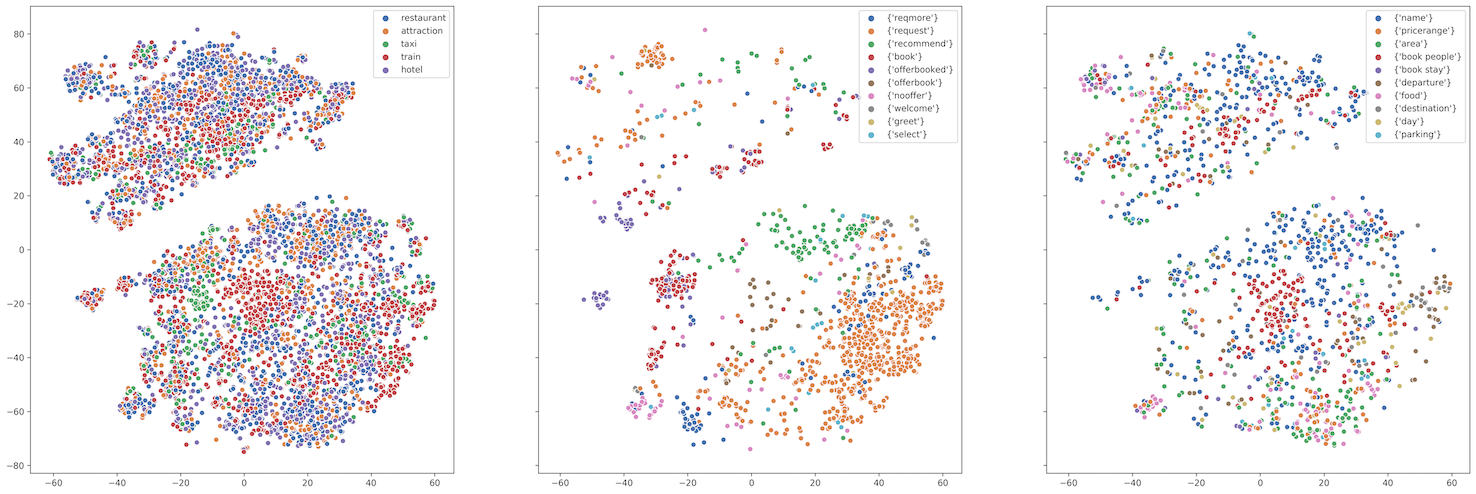}
\caption{The tSNE visualization of dialogue representations from the ELECTRA-Dis. (Best view in color.)}
\label{fig:tsne-electragen}
\end{figure*}

\begin{figure*}[t]
\centering
\includegraphics[width=\linewidth]{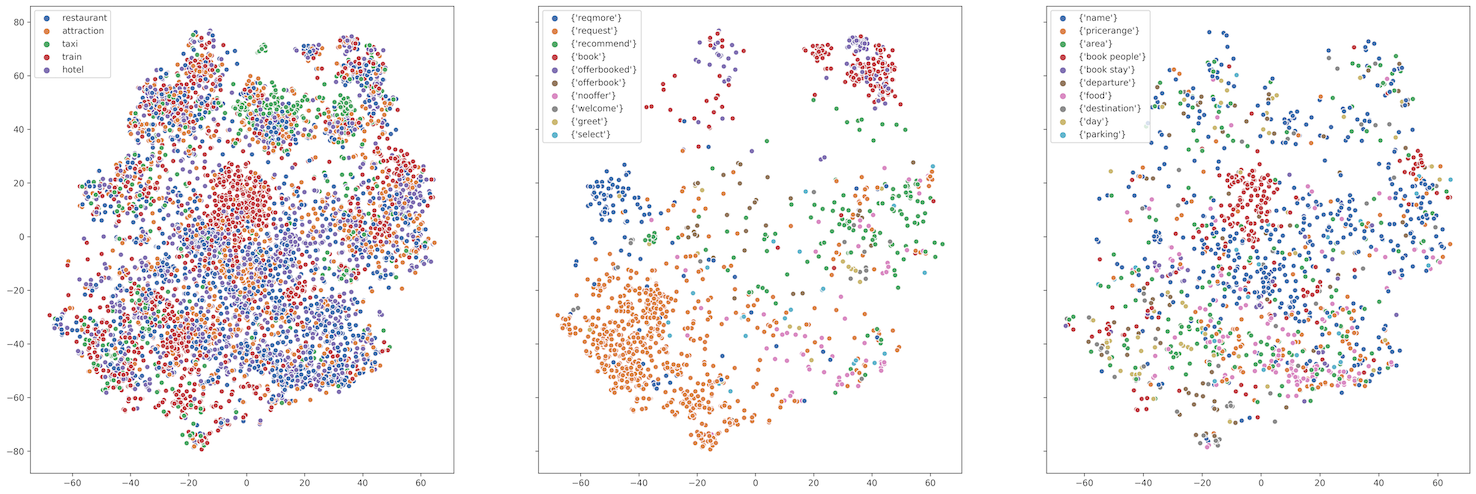}
\caption{The tSNE visualization of dialogue representations from the ELECTRA-Dis. (Best view in color.)}
\label{fig:tsne-electradis}
\end{figure*}

\begin{figure*}[t]
\centering
\includegraphics[width=\linewidth]{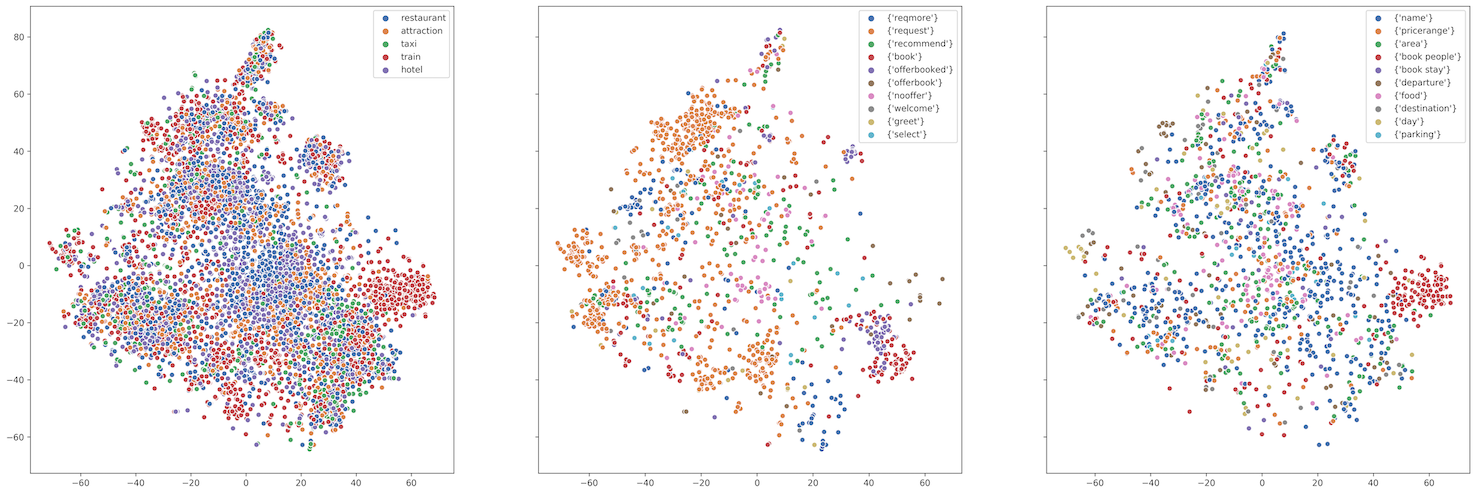}
\caption{The tSNE visualization of dialogue representations from the RoBERTa. (Best view in color.)}
\label{fig:tsne-roberta}
\end{figure*}

\begin{figure*}[t]
\centering
\includegraphics[width=\linewidth]{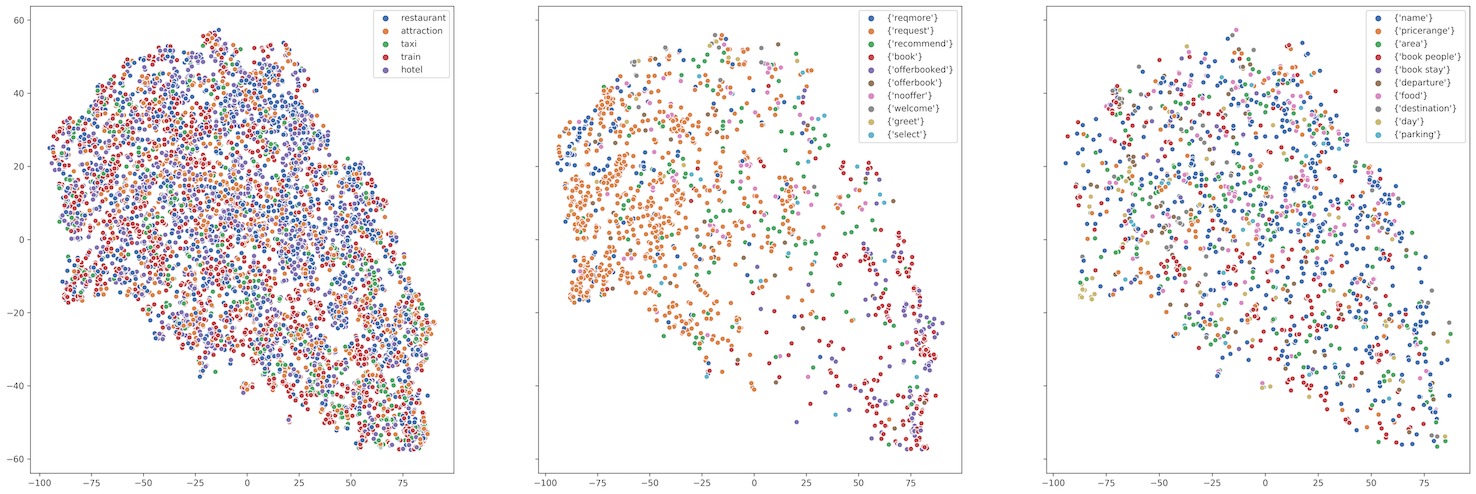}
\caption{The tSNE visualization of dialogue representations from the DialoGPT. (Best view in color.)}
\label{fig:tsne-dialogpt}
\end{figure*}

\begin{figure*}[t]
\centering
\includegraphics[width=0.9\linewidth]{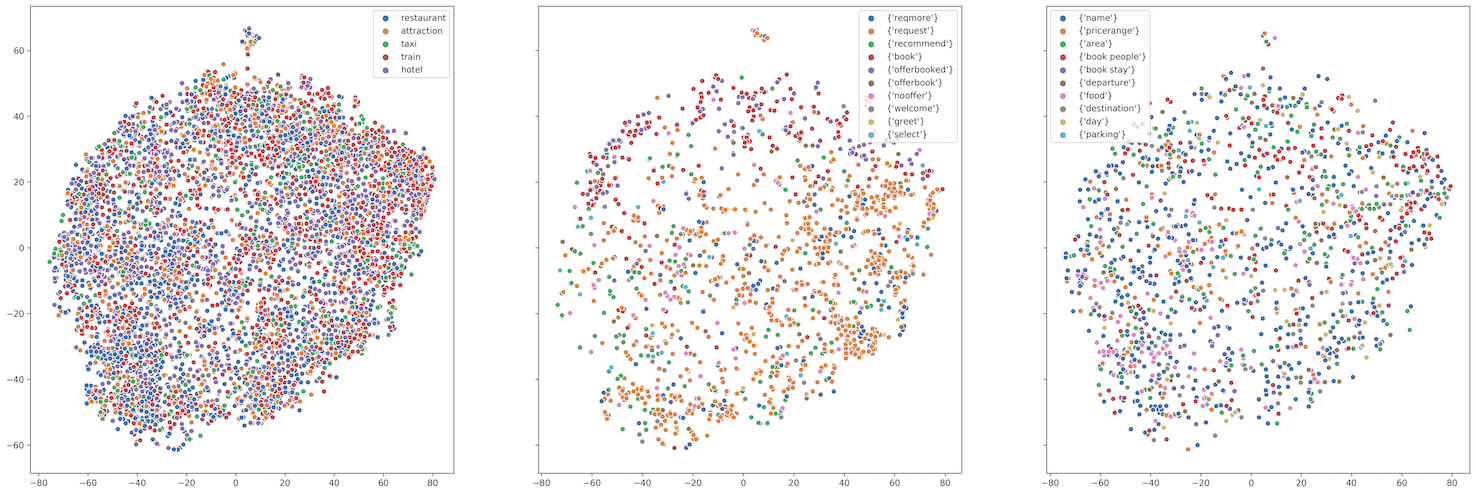}
\caption{The tSNE visualization of dialogue representations from the AlBERT. (Best view in color.)}
\label{fig:tsne-albert}
\end{figure*}

\begin{table*}[t]
\centering
\resizebox{0.8\linewidth}{!}{
\begin{tabular}{|c|l|}
\hline
\multicolumn{2}{|c|}{\textbf{TOD-BERT-jnt}} \\ \hline
\multirow{5}{*}{\begin{tabular}[c]{@{}c@{}}Cluster 1\\ (Restaurant Request)\end{tabular}} 
& i have many options available for you ! is there a certain area or cuisine that interests you ? \\ \cline{2-2} 
& there are 21 restaurant -s available in the centre of town . how about a specific type of cuisine ? \\ \cline{2-2} 
& do you have any specific type of food you would like ? \\ \cline{2-2}  
& there 33 place -s that fit your criteria . do you have a particular cuisine type in mind so that i can narrow the results down ? \\ \cline{2-2} 
& is there a particular cuisine you are looking for ? \\ \hline

\multirow{5}{*}{\begin{tabular}[c]{@{}c@{}}Cluster 2\\ (Taxi/Train)\end{tabular}} 
& what time do you want to leave and what time do you want to arrive by ? \\ \cline{2-2} 
& do you have a time preference ? \\ \cline{2-2} 
& when would you like to leave and arrive ? \\ \cline{2-2}  
& what time would you like to leave the junction ? \\ \cline{2-2} 
& wonderful , i can help you . what time on sunday would you like to depart ? \\ \hline

\multirow{5}{*}{\begin{tabular}[c]{@{}c@{}}Cluster 3\\ (Attraction Recommend)\end{tabular}} 
& i can recommend the allenbell . it s in the east , is cheap yet has a 4 star rating and free wifi and parking . can i help you book ? \\ \cline{2-2} 
& the university arms is an expensive , 4 star hotel with free wifi . comparatively , the alexander bed and breakfast is a cheap -ly priced guesthouse , also 4 stars . \\ \cline{2-2} 
& i have found the guesthouse you were wanting . would you like me to book this for you ? \\ \cline{2-2}  
&  how about the express by holiday inn cambridge , it s in the east . \\ \cline{2-2} 
&  the expensive 1 is actually not much more than the other 2 . i would highly recommend it . that would be at the express by holiday inn cambridge . it s in the east . \\ \hline

\multirow{5}{*}{\begin{tabular}[c]{@{}c@{}}Cluster 4\\ (Hotel Inform)\end{tabular}} 
& the address is hills road city centre \\ \cline{2-2} 
& their address is unit g6 , cambridge leisure park , clifton road . the postcode is cb17dy . \\ \cline{2-2} 
& the address is corn exchange street . is there anything else i can help you with ? \\ \cline{2-2}  
& yes , the phone number is 01223277977 . the address is hotel felix whitehouse lane huntingdon road , and the post code is cb30lx . want to book ? \\ \cline{2-2} 
& the bridge guest house is at 151 hills road and their number is 01223247942 . \\ \hline

\multirow{5}{*}{\begin{tabular}[c]{@{}c@{}}Cluster 5\\ (Welcome/End)\end{tabular}}
& you are welcome . is there anything else i can help you with today ? \\ \cline{2-2} 
& great . is there anything else that you need help with ? \\ \cline{2-2} 
& is there anything else that you would like ? \\ \cline{2-2}  
& no problem . can i help you with anything else ? \\ \cline{2-2} 
& is there something else i can help you with then ? \\ \hline
\end{tabular}
}
\caption{Clustering results of the TOD-BERT-jnt model. The samples are randomly picked from each randomly selected five clusters (using K=32).}
\label{tab:todbert}
\end{table*}

\begin{table*}[t]
\centering
\resizebox{0.8\linewidth}{!}{
\begin{tabular}{|c|l|}
\hline
\multicolumn{2}{|c|}{\textbf{GPT2}} \\ \hline
\multirow{5}{*}{\begin{tabular}[c]{@{}c@{}}Cluster 1\\ \end{tabular}} 
& there are 9 indian restaurant -s in centre what price range do you want ? \\ \cline{2-2} 
& do you have any specific type of food you would like ? \\ \cline{2-2} 
& 105 minutes is the total travel time . can i help you with anything else ? \\ \cline{2-2}  
& there are lots to choose from under that criteria . what day would you like to travel on ? \\ \cline{2-2} 
& i have the cote in the centre . it is in the expensive range . would you like to make a booking ? \\ \hline

\multirow{5}{*}{\begin{tabular}[c]{@{}c@{}}Cluster 2\\\end{tabular}} 
& your reference number is x5ny66zv . \\ \cline{2-2} 
& i booked tr3932 , reference number is fiw5abo2 . \\ \cline{2-2} 
& nusha is in the south , and the phone number is 01223902158 . \\ \cline{2-2}  
& they are located at 12 lensfield road city centre , postcode cb21eg , and phone number 01842753771 . \\ \cline{2-2} 
& it would cost 16.50 pounds . \\ \hline

\multirow{5}{*}{\begin{tabular}[c]{@{}c@{}}Cluster 3\\\end{tabular}} 
& i hope i have been of help \\ \cline{2-2} 
& the entrance fee is free . anything else i can do for you today ? \\ \cline{2-2} 
& sure , lookout for a blue volvo the contact number is 07941424083 . can i help with anything else ? \\ \cline{2-2}  
&  1 moment while i try to make the reservation of table for 8 , friday at 16:30 . \\ \cline{2-2} 
&  i have 3 options for you 2 in the north in the moderate price range and 1 that s expensive in the east . \\ \hline

\multirow{5}{*}{\begin{tabular}[c]{@{}c@{}}Cluster 4\\ \end{tabular}} 
& when would you like to leave and arrive ? \\ \cline{2-2} 
& booking was unsuccessful . can you try another time slot ? \\ \cline{2-2} 
& on what day will you be traveling ? \\ \cline{2-2}  
& tr3823 will arrive at 16:55 , would that work for you ? \\ \cline{2-2} 
& okay , what day did you have in mind ? \\ \hline

\multirow{5}{*}{\begin{tabular}[c]{@{}c@{}}Cluster 5\\ \end{tabular}} 
& saffron brasserie is an expensive restaurant that serves italian food \\ \cline{2-2} 
& there are 21 restaurant -s available in the centre of town . how about a specific type of cuisine ? \\ \cline{2-2} 
& i have 5 different restaurant -s to choose from . there are 4 in the centre of town , and 1 in the west . do you have a preference ? \\ \cline{2-2}  
& i have about 5 different entertainment venue -s if that is what you are looking for . do you have a preference on the area its located in ? \\ \cline{2-2} 
& there are no colleges close to the area you are requesting , would you like to chose another destination ? \\ \hline
\end{tabular}
}
\caption{Clustering results of the GPT2 model. The samples are randomly picked from each randomly selected five clusters (using K=32).}
\label{tab:gpt}
\end{table*}

\begin{table*}[t]
\centering
\resizebox{0.8\linewidth}{!}{
\begin{tabular}{|c|l|}
\hline
\multicolumn{2}{|c|}{\textbf{DialoGPT}} \\ \hline
\multirow{5}{*}{\begin{tabular}[c]{@{}c@{}}Cluster 1\\ \end{tabular}} 
& it is located in jesus lane \\ \cline{2-2} 
& your booking was successful , the reference number is waeyaq0m . may i assist you with anything else today ? \\ \cline{2-2} 
& your booking is successful ! your reference number is iigra0mi . do you need anything else ? \\ \cline{2-2}  
& 1 moment while i try to make the reservation of table for 8 , friday at 16:30 . \\ \cline{2-2} 
& this booking is successful for 1 night . your reference number is 85bgkwo4 . is there anything else i can assist you with ? \\ \hline

\multirow{5}{*}{\begin{tabular}[c]{@{}c@{}}Cluster 2\\\end{tabular}} 
& sure , how many days and how many people ? \\ \cline{2-2} 
& i recommend castle galleries and it s free to get in ! \\ \cline{2-2} 
& i have plenty of trains departing from leicester , what destination did you have in mind ? \\ \cline{2-2}  
& i have 5 colleges in the centre area . what specific college are you looking for ? \\ \cline{2-2} 
& oh yes quite a few . which part of town will you be dining in ? \\ \hline

\multirow{5}{*}{\begin{tabular}[c]{@{}c@{}}Cluster 3\\\end{tabular}} 
& i have many options available for you ! is there a certain area or cuisine that interests you ? \\ \cline{2-2} 
& there are lots to choose from under that criteria . what day would you like to travel on ? \\ \cline{2-2} 
& actually all 5 have free wifi . what star rating would you like ? \\ \cline{2-2}
&  i have found the guesthouse you were wanting . would you like me to book this for you ? \\ \cline{2-2} 
& yes , the hamilton lodge has internet . \\ \hline

\multirow{5}{*}{\begin{tabular}[c]{@{}c@{}}Cluster 4\\ \end{tabular}} 
& its entrance fee is free . \\ \cline{2-2} 
& sure , lookout for a blue volvo the contact number is 07941424083 . can i help with anything else ? \\ \cline{2-2}  
& how many people is the reservation for ? \\ \cline{2-2} 
& how about train tr3934 ? it leaves at 12:34 and arrives at 13:24 . travel time is 50 minutes . \\ \cline{2-2}
& sure , the phone number is 01223902112 and they are in postcode cb58sx . can i help you with anything else today ? \\ \hline

\multirow{5}{*}{\begin{tabular}[c]{@{}c@{}}Cluster 5\\ \end{tabular}} 
& yes i can . what restaurant are you looking for ? \\ \cline{2-2} 
& what time would you like to leave the junction ? \\ \cline{2-2} 
& no problem . can i help you with anything else ? \\ \cline{2-2}  
& you are welcome . is there anything else i can help you with today ? \\ \cline{2-2} 
& is there anything else i can help you with ? \\ \hline
\end{tabular}
}
\caption{Clustering results of the DialoGPT model. The samples are randomly picked from each randomly selected five clusters (using K=32).}
\label{tab:dialogpt}
\end{table*}



\end{document}